%% file: main.tex
\documentclass{article}

\usepackage[preprint]{neurips_2022}

\usepackage[utf8]{inputenc} % allow utf-8 input
\usepackage[T1]{fontenc}    % use 8-bit T1 fonts
\usepackage{hyperref}       % hyperlinks
\usepackage{url}            % simple URL typesetting
\usepackage{booktabs}       % professional-quality tables
\usepackage{amsfonts}       % blackboard math symbols
\usepackage{nicefrac}       % compact symbols for 1/2, etc.
\usepackage{microtype}      % microtypography
\usepackage{xcolor}         % colors
\usepackage{stmaryrd}
\usepackage{algorithm, algpseudocode}
\usepackage{mwe}
\usepackage{multirow}

\input{math_commands.tex}

\usepackage{hyperref}
\hypersetup{
    colorlinks=true,
    linkcolor=red,
    filecolor=magenta,
    urlcolor=blue,
    citecolor=purple,
    pdftitle={Overleaf Example},
    pdfpagemode=FullScreen,
    }

\usepackage{graphicx}
\usepackage{caption}
\usepackage{comment}
\usepackage{amsmath}
\usepackage{wrapfig}

\title{On making optimal transport robust to all outliers}

\author{%
  Kilian Fatras\\
  Mila - Quebec AI Institute, McGill University\\
  Montr\'eal, Qu\'ebec, Canada \\
  \texttt{kilian.fatras@mila.quebec} \\
}

\begin{document}

\maketitle

%\begin{abstract}
%Optimal transport (OT) seeks the minimum displacement cost of moving a measure to another. To transport measures, OT relies on marginal constraints which makes it sensitive to outliers present in the measures. Outlier robust OT variants have been proposed based on the definition that outliers are samples which are expensive to move to the other distribution. In this paper, we show that this is a restricted definition as for instance, outliers can be noisy samples which are close to the other distribution and far from clean samples.  Outlier robust OT variants fail to address this problem and they even aggravate the problem by transporting these samples more. To make them robust against such outliers, we propose to train a classifier to classify source and target samples, where outliers can be seen as noisy label samples in this context. Based on the difference between the classification and the given label, we then propose two reweighting strategies to mitigate the weights of the detected outliers. We evaluate our method on Wasserstein barycenter, gradient flow, generative models and label propagation experiments which show that we successfully detected the outliers and did not consider them in the transport problem.
%\end{abstract}

\begin{abstract}
    Optimal transport (OT) is known to be sensitive against outliers because of its marginal constraints. Outlier robust OT variants have been proposed based on the definition that outliers are samples which are expensive to move. In this paper, we show that this definition is restricted by considering the case where outliers are closer to the target measure than clean samples. We show that outlier robust OT fully transports these outliers leading to poor performances in practice. To tackle these outliers, we propose to detect them by relying on a classifier trained with adversarial training to classify source and target samples. A sample is then considered as an outlier if the prediction from the classifier is different from its assigned label. To decrease the influence of these outliers in the transport problem, we propose to either remove them from the problem or to increase the cost of moving them by using the classifier prediction. We show that we successfully detect these outliers and that they do not influence the transport problem on several experiments such as gradient flows, generative models and label propagation.
\end{abstract}

\section{Introduction}\label{sec:intro}

%\paragraph{Computing distances between probability measures with OT}
Comparing probability measures is a fundamental problem in machine learning, especially as many probability measures have disjoint supports. One can use Optimal Transport (OT) \citep{COT_Peyre} to achieve this. OT compares probability measures by looking for the minimum displacement cost from a measure to another with respect to a ground cost. To ensure that all mass is transported, the marginals of the optimal transport plan need to be equal to the input probability measures. When the ground cost is a distance, OT defines a distance between probability measures, known as the Wasserstein distance (denoted $W$). This ability to discriminate measures has made it a useful tool for many machine learning applications. It has been used in domain adaptation to transport source labeled samples to the target domain \citep{Courty_OTDA} or to align samples in an embedded space in \cite{courty_jdot, Damodaran_2018_ECCV, fatras21a}. OT has also been used as a loss function in supervised learning \citep{frogner_2015} and in generative models \citep{arjovsky17a, genevay18a, burnel2021}. Despite its many successes, OT has been shown to be sensitive to outliers. Indeed when datasets are tainted by outliers, the marginal constraints on the transport plan force OT to transport all samples including outliers. In practice, it impacts negatively the performances of neural networks, as for instance in generative models, where the generators would generate outliers \citep{balaji2020robust}.% or in partial domain adaptation \citep{fatras21a} where some source classes would be transported to the target domain where they are not present.

%\paragraph{OT robust to outliers}
This weakness to outliers raised some attention in the machine learning community and several optimal transport variants were proposed to fix it, \emph{i.e., variants which would not transport outliers}. The main considered direction to make OT robust to outliers was to decrease the transported mass of outliers. To do so, they replaced the hard marginal constraints by soft penalties following the framework of unbalanced OT (UOT) \citep{Liero_2017, Hanin1992, piccoli2014}. This formulation considers what is the more costly between transporting the samples or violating the mass penalty. Built upon this theory, \citep{balaji2020robust, RobustOptimalTransport2022Nietert} looked for relaxed measures which minimized the optimal transport cost.
Another workaround was to add a Total Variation norm term in the objective which was shown to correspond to a truncated ground cost  \citep{mukherjee2020outlierrobust}. However, their definition of outliers is restricted and we highlight some weaknesses.

Outliers were defined in previous work \citep{balaji2020robust, mukherjee2020outlierrobust, RobustOptimalTransport2022Nietert} as \emph{samples which are costly to transport} and we refer to them as outliers of \emph{first type}, formally:

\begin{definition}\label{def:outlier_ot}
Let $\alpha, \beta$ be two probability measures $\alpha, \beta \in \mathcal{M}_+^1(\mathcal{X})$, where $\mathcal{M}_{+}^1(\mathcal{X})$ is the set of probability measures on $\mathcal{X}$. We consider that $\alpha$ is the weighted sum of two other probability measures, formally let $\alpha = (1-\kappa) \alpha_c + \kappa \alpha_o$, where $\alpha_c$ is the clean sample measure, $\alpha_o$ the outlier measure and $\kappa \in [0,1]$. Then, $\alpha_o$ is considered a first type (or type I) outlier measure if $W(\alpha_c, \beta) \leq W(\alpha_o, \beta)$.
\end{definition}
%\paragraph{Outliers and ground cost}
%The robust variants to outliers are then well designed to handle the case where $W((1-\kappa)\alpha_c + \kappa \delta_{\zz_{0}}, \beta) \rightarrow \infty$ when $c(\zz_{0}, \yy) \rightarrow \infty$ for all $\yy$.

%The first weakness of this definition is that it relies on the ground cost to define outliers. So a sample could be considered as an outlier for some distances and as a clean sample for others. Furthermore,
This restricted definition of outliers leads to a problem which has not been discussed in the literature before, where a second type of outliers are always transported as opposed to what we want. The case where \emph{outliers are closer to the target measure than the majority of clean samples}. For instance, suppose we want to transport a source dataset of wolf images to a target dataset of dog images with noisy labels. In a label noise scenario, some dog images could be labeled as wolf leading to wolf outliers close to the measure of dog images and vice-versa. This scenario corresponds to a mixture of $\alpha_c$ and $\beta$, $\alpha = (1-\kappa) \alpha_c + \kappa \beta$. We refer to these outliers as \emph{second type outliers} and define them as:

\begin{definition}\label{def:outlier_ot2}
By considering the same notations as in Definition \ref{def:outlier_ot}. $\alpha_o$ is considered as a second type (or type II) outlier measure if $W(\alpha_o, \beta) < W(\alpha_c, \beta)$.
\end{definition}

\begin{figure}[t]
  \begin{center}
    \includegraphics[width=1.\linewidth]{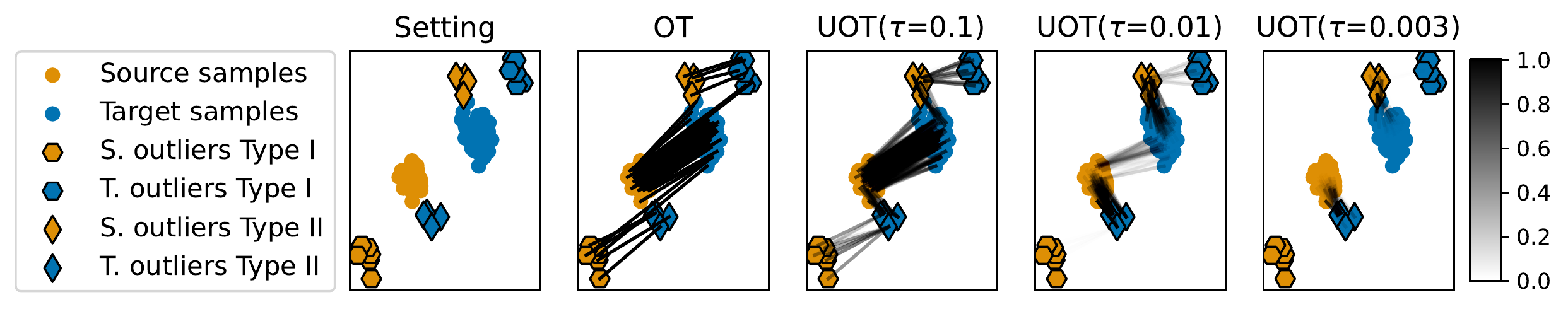}
  \end{center}
  \vspace{-0.2cm}
    \caption{(Best viewed in color) Illustration of exact and unbalanced optimal transport plan on a 2D dataset for different marginal penalization $\tau$. The source and target measures are tainted with outliers of type I and II. Intensities are normalized by the maximum value of the transport plan.}
    \label{fig:toy_uot_outlier}
  \vspace{-0.3cm}
\end{figure}

We illustrate the weakness of UOT based losses to outliers of type II in Figure \ref{fig:toy_uot_outlier}. We show that for small values of the marginal penalization constant $\tau$, UOT only transports outliers of type II. We consider two uniform 2D probability measures composed of 75 samples each with 10 outliers. 6 outliers of type I are far from both measures while the 4 remaining are of type II and closer to the other measure than the clean samples. We plot the connections between samples and normalized the connection intensities by the largest intensity of the OT plan. We can even observe that OT, which corresponds to UOT as $\tau \rightarrow +\infty$, transports all samples including outliers, while by choosing smaller $\tau$ values, UOT does not transport outliers of type I anymore. While UOT has successfully improved robustness to type I outliers, its formulation has in fact aggravated the sensibility to the second type. Indeed, due to the small ground cost values, UOT assigns to second type outliers a greater transport weight than to clean samples which is the opposite of what we would like. These samples would not increase the OT cost but can lead to poor performances in the context of data fitting problem (see Section \ref{sec:experiments}). As robust OT variants are based on UOT, they also fail to tackle second type outliers.
%In the worst case scenario for very small $\tau$ values, the biggest and few connections are the connections of outliers close to the other measures. This is what we do not want as we transport outliers instead of clean samples. In this setting, all robust formulation would transport these outliers because they are not costly to transport with respect to clean samples.

%Finally, a sample could be considered as an outlier for some distances and as a clean sample for others.

In this paper, our contributions are the following. After showing that robust OT variants are sensitive to outliers of second type, we propose a method to detect these outliers by considering a classification problem on source and target samples. In this scenario, second type outliers can be seen as noisy label samples and we show that classifiers overfit over them. Thus, we want our classifier to be robust against noisy labels and to do this, we use the virtual adversarial training algorithm \citep{Miyato2019} which was shown to be robust against noisy labels \citep{Fatras2021WAR}. When the classification of a sample does not match its assigned label, we consider the sample as an outlier. We then decrease the influence of second type outliers by using a hard reweighting strategy, \emph{i.e.,} by assigning them a weight equal to 0, or a soft reweighting strategy, \emph{i.e.,} by modifying the used ground cost to take into account the prediction of the classifier to increase the cost of moving them.

%\begin{figure}[t]
%    \centering
%        \includegraphics[width=0.8\linewidth]{figs/ot_plans_uot.pdf}
%    \caption{Illustration of unbalanced optimal transport plan on a 2D dataset for different $\tau$ coefficients. The source and target distributions are tainted with outliers which are close or far to the other distribution.}
%    \label{fig:toy_uot_outlier}
%\end{figure}

%\paragraph{Structure of the paper}
The paper is structured as follows. In Section \ref{sec:rw}, we discuss the different OT variants robust to outliers and we review some definitions and methods of outlier detection. In Section \ref{sec:proposed_methods}, we present our method to mitigate the effect of second type outliers as well as the components it is built upon such as the virtual adversarial training. In Section \ref{sec:experiments}, we empirically evaluate our method on gradient flow, generative adversarial networks (GANs) and label propagation experiments.

\section{Related work on robust optimal transport and outliers}\label{sec:rw}

In this section, we define formally the optimal transport cost as well as some robust OT variants and then we discuss standard methods to detect outliers.

%\subsection{Optimal transport and robust variants}

\paragraph{Optimal Transport} The optimal transport cost measures the minimal displacement cost of moving a probability measure $\alpha$ to another probability measure $\beta$ with respect to a ground metric $c$ on the data space $\mathcal{X}$ and $\mathcal{Z}$ \citep{COT_Peyre}. Formally, let $(\alpha, \beta) \in \mathcal{M}_{+}^1(\mathcal{X}) \times \mathcal{M}_{+}^1(\mathcal{Z})$. For a ground cost on the data space $c:\mathcal{X}\times \mathcal{Z} \mapsto \mathbb{R}_+$, the Kantorovich primal problem between the two measures $\alpha, \beta$ is
\begin{equation}
    W_{c}(\alpha, \beta) = \underset{\pi \in \mathcal{U}(\alpha, \beta)}{\text{min}} \int_{\mathcal{X}\times \mathcal{Z}}c(\xx,\zz) d\pi(\xx,\zz),
\label{eq:wasserstein_dist}
\end{equation}
where $\mathcal{U}(\alpha, \beta)$ is the set of joint probability measures with
marginals $\alpha$ and $\beta$ such that
$
\mathcal{U}(\alpha, \beta) = \left \{ \pi \in \mathcal{M}_{+}^1(\mathcal{X}, \mathcal{Z}): \PP_{\mathcal{X}}\#\pi = \alpha, \PP_{\mathcal{Z}}\#\pi = \beta \right\}\nonumber
$. {$\PP_{\mathcal{X}}\#\pi$ (resp.\ $\PP_{\mathcal{Z}}\#\pi$) is the marginalization of $\pi$ over $\mathcal{X}$ (resp.\ $\mathcal{Z}$)}. These constraints enforce that all the mass from $\alpha$ is transported to $\beta$ and vice-versa, which makes this formulation sensitive to outliers. %The problem of this formulation, named Kantorovich primal problem, is that marginal constraints makes OT sensitive to outliers as they would be transported.

\paragraph{Unbalanced Optimal Transport} %While first used to compare measures of different mass by relaxing the conservation of mass constraints, the unbalanced optimal transport cost has also been used for its robustness to outlier properties.
Unbalanced optimal transport was initially used to compare measures of different mass by relaxing the conservation of mass constraints. It replaces the OT hard marginal constraints by soft penalties using Csisz\`ar divergences \citep{Liero_2017, figalli_partial, chapel2020partial} or integral probability metrics~\citep{nath2020unbalanced}. This relaxation has made UOT appealing in the context of robustness against outliers. Formally, consider the Kullback-Leibler divergence ($ \texttt{KL}(\xx\|\yy) = \sum_{i} \xx_{i}\log(\frac{\xx_{i}}{\yy_i}) - \xx_{i} + \yy_i$) and two positive measures $\alpha, \beta \in \mathcal{M}_{+}(\mathcal{X})$, the entropic UOT program between measures with cost $c$ is defined as
\begin{equation}
  \operatorname{OT}_{\texttt{KL}}^{\tau, \varepsilon}(\alpha, \beta, c) = \underset{\pi \in \mathcal{M}_+(\mathcal{X} \times \mathcal{Z})}{\text{min}} \int cd\pi + \varepsilon \texttt{KL}(\pi\|\alpha \otimes \beta)  + \tau (\texttt{KL}(\pi_1\|\alpha)  + \texttt{KL}(\pi_2\|\beta)),
\label{eq:uot_def}
\end{equation}
where $\pi$ is the transport plan, $\pi_1$ and $\pi_2$ the plan's marginals, $\tau$ is the marginal penalization and $\varepsilon \geq 0$ is the entropic regularization coefficient \citep{CuturiSinkhorn}. The solution is computed via a generalized Sinkhorn algorithm~\citep{ChizatPSV18} with a complexity of $\tilde{O}(n^2/\epsilon)$~\citep{pham20a}. Note that the marginals of $\pi$ are no longer equal to $(\alpha,\beta)$ in general and that the \texttt{KL} divergence can be replaced by the TV-norm $\texttt{TV}$ to define partial OT \citep{figalli_partial}. UOT showed robustness properties against Dirac outliers \citep{fatras21a} which explains why it is at the heart of robust OT variants.

\paragraph{Robust Optimal Transport} The robust optimal transport \citep{RobustOptimalTransport2022Nietert} is defined as
\begin{equation}
\operatorname{ROT}_\rho(\alpha, \beta, c) =  \min_{\substack{\alpha^\prime
\beta^\prime \in \mathcal{M}_+(\mathcal{X}); \\ \alpha^\prime \leq \alpha, \|\alpha^\prime - \alpha\|_{\texttt{TV}} \leq \rho; \\ \beta^\prime \leq \beta, \|\beta^\prime - \beta\|_{\texttt{TV}} \leq \rho}} \operatorname{W}_{c}\left(\frac{\alpha^\prime}{\alpha^\prime(\mathcal{X})}, \frac{\beta^\prime}{\beta^\prime(\mathcal{X})} \right),
\label{eq:robust}
\end{equation}

which is similar to \citep{balaji2020robust} except that they do not require $\alpha^\prime, \beta^\prime$ to be probability measures. The intuition is that by minimizing the OT cost, outliers, which are costly to transport, would have a smaller mass. To prevent over-relaxation, they control the distances between the relaxed and original probability measures with a $f$-divergence such as TV-norm. \cite{mukherjee2020outlierrobust} used a truncated ground cost $c_{\lambda}$ in the original OT formulation (Eq. \ref{eq:wasserstein_dist}) and thus, their formulation can be solved with any modern OT solver. They solved $W_{c_{\lambda}}(\alpha, \beta)$ with $c_{\lambda}(\xx_i, \yy_j) = \min\{c(\xx_i, \yy_j), 2\lambda \}$.
Another direction was proposed in \citep{staerman21a} where authors used a median of means approach to tackle the dual Kantorovich problem from a robust statistics perspective. The current method to detect outliers is to use the value of the ground cost and to tune a given hyperparameter to control the marginal penalization and avoid transporting costly to move samples.  In the next paragraph, we discuss some other definitions and methods to detect outliers not related to the OT literature.

\paragraph{Detection of outliers}

Many outlier definitions have been proposed in the literature but none of them has been universally accepted. For instance \cite{barnett1994outliers} defines an outlier to be one or more observations which are not consistent among others, while \cite{Navarro2021} defines an outlier as a sample outside a certain ball centered on a considered sample. To detect outliers, several methods try to estimate the covariance matrix of samples \citep{Pena2001, Friedman1987, MinimumCovarianceDeterminant1999}. Other methods rely on clustering algorithms \citep{Ester96, Guha98, knorr98} or random forest \citep{Liu2008} to isolate and detect outliers. Another strategy is to learn the border of the empirical measure using one class SVM \citep{Scholkopf1996}. Lastly, \cite{Breunig2000} developed a score measuring the local density deviation of a given sample with respect to its neighbors to detect outliers. Some surveys on outlier detection can be found in \cite{Aggarwal2017, Pang2021}. A close notion of outliers considered in this work comes from supervised learning with noisy label, where samples with noisy labels can be seen as outliers. The main learning strategies against label noise are based on data cleaning methods \citep{Brodley1999,Vahdat17}, transition probability-based methods \citep{Liu_2014, Patrini_2017_CVPR} or regularization methods \citep{Li2020DivideMix, NEURIPS2018_a19744e2}.

%We present our method to detect outliers in the next section.

\section{Proposed methods}\label{sec:proposed_methods}

In this section, we present our algorithm to detect the presence of second type outliers, and we present how we modify existing robust to outliers OT costs to tackle them.

%\paragraph{Considered outliers}
%In the OT literature, outliers are usually defined as samples costly to move to the other distribution (see Definition \ref{def:outlier_ot}). We illustrated that it is a restricted definition in Section \ref{sec:intro} as outliers can be closer to the other distribution. Thus we consider a definition closer to the definition used in \cite{barnett1994outliers} \emph{an outlier are observations, which are not consistent among others in their distribution}. While costly to move samples are handled thanks to the current robust OT variants, we present in the next paragraphs how we handle the second type of outliers.

\paragraph{Classification problem and adversarial training}
Our purpose is to make robust OT variants robust to second type outliers as they are already robust to first type outliers. To this end, we propose to learn a classifier to distinguish samples depending on whether they have been drawn from the measure $\alpha$ or $\beta$. The classifier would learn a boundary between the two measures $\alpha$ and $\beta$ and could be used to detect outliers similarly to the one class SVM \citep{Scholkopf1996}. Misclassified samples, samples drawn from a measure but predicted to belong to the other, would be considered as outliers and we would only rely on correctly classified samples. Our setup is a binary classification problem on source and target samples with a deep neural network (DNN), where we assign to samples a label illustrating their assigned measure, \emph{i.e.,} $\yy_s$ (resp. $\yy_t$) is the source (resp. target) label.%formally our empirical measures are $\alpha = \sum_{i=1}^n \delta_{\xx_i, \yy_s}$ and $\beta = \sum_{j=1}^n \delta_{\zz_j, \yy_t}$,

In the context of a dataset tainted by outliers, the second type outliers can be seen as noisy label samples. It is well known that DNNs trained with cross-entropy (denoted $\text{CE}$) have strong memory abilities and can easily overfit on noisy labels \citep{Zhang_2017}, which is the case in our experiments (see Section \ref{sec:experiments}). To prevent this overfitting issue we want to smooth the decision boundaries of the classifier where it is locally broken because of noisy labels. This can be done using adversarial training where the prediction of perturbed inputs are forced to match the label of the clean input. We look for the solution of
\begin{equation}\label{eq:robust}
\argmin_\theta \sum_{i=1}^N \max_{\xx^u_i \in \mathcal{V}_i} L(f_{\theta}(\xx_i^u),\yy_i),
\end{equation}

where $\mathcal{V}_i$ is the neighborhood of input $\xx_i$. Unfortunately, the perturbation $\xx^u_i$ is intractable in practice and that is why we rely on adversarial training \citep{Goodfellow2015} where the max is replaced by the direction which produces the maximum variation in the prediction. However adversarial training uses labels which can be potentially noisy, thus we rely on the adversarial regularization which replaces the label used in adversarial training by the classifier prediction \cite{Miyato2019}. Let $\omega=0.001$, the optimized loss is then
\begin{align}
    &L_{\text{AR}}(f_{\theta}(\xx_i),\yy_i) = \omega\text{CE}(f_{\theta}(\xx_i),\yy_i) + \operatorname{R}_{\texttt{KL}}(\xx_i,{f_\theta}),\\\label{eq:AR}
    &\quad \text{with } \operatorname{R}_{\texttt{KL}}(\xx_i,{f_\theta}) = \texttt{KL}(f_{\theta}(\xx_i+ \rr_i^a)\| f_{\theta}(\xx_i)),\;\; \nonumber \\
    &\quad \text{and } \rr_i^a = \underset{\rr_i,\|\rr_i\| \leq \eta}{\text{argmax }} \texttt{KL}(f_{\theta}(\xx_i + \rr_i)\|f_{\theta}(\xx_i)),
\end{align}
where $f_{\theta}(\xx_i + \rr^a)$ is the neural network prediction of the adversarial input. $\operatorname{R}_{\texttt{KL}}$ forces the local Lipschitz constant to be small with respect to the $\texttt{KL}$ divergence. The adversarial regularization was first designed to solve semi-supervised learning problem \citep{Miyato2019}, but it has been proven to be efficient against label noise as the classifier learns over an interpolation of the label and the prediction \citep{Fatras2021WAR}. The adversarial direction is approximated using the power iteration algorithm \citep{GOLUB2000}. Thus we want to take advantage of this regularization to detect samples which are annotated from a measure but classified as another like second type outliers. In the next paragraph, we describe how we use the classifier output to mitigate their influence.

\begin{figure}[t!]
  \begin{center}
    \includegraphics[width=0.7\linewidth]{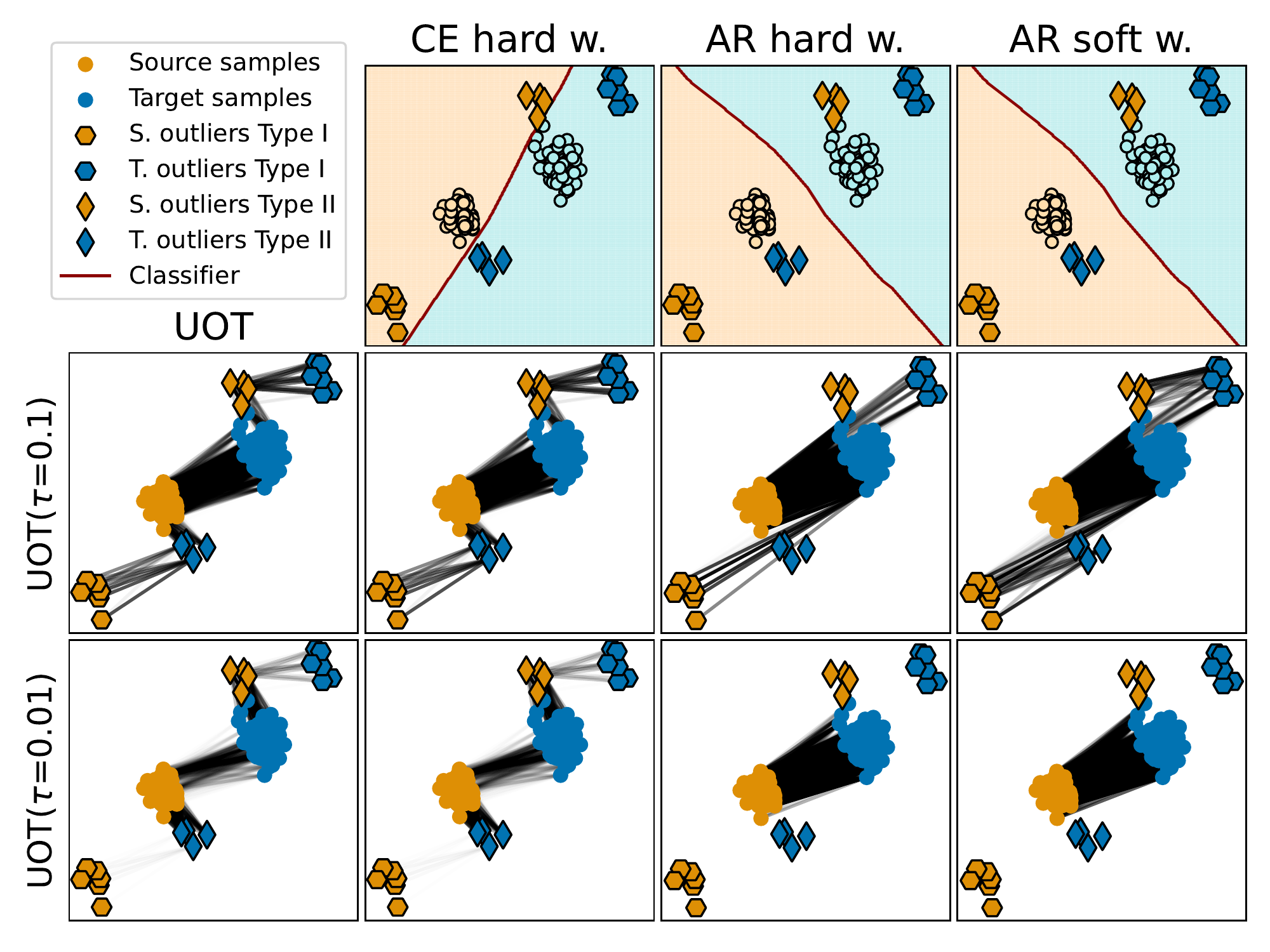}
  \end{center}
  \vspace{-0.3cm}
    \caption{(Best viewed in color) Illustration of our different strategies with unbalanced optimal transport plan on a 2D dataset for different marginal penalization $\tau$. (First row) classification boundary for different classifiers. (Second and third row) UOT transport plan for different strategies and different $\tau$. The source and target measures are tainted with first (6 samples) and second (4 samples) type outliers. Intensities are normalized by the maximum value of the transport plan.}
    \label{fig:toy_classif_outlier}
  \vspace{-0.3cm}
\end{figure}

\paragraph{Mitigating influence of outliers}
Once outliers have been detected using our classifiers, we can define a \emph{hard reweighting strategy} to decrease their weight in the empirical measures. It corresponds to assigning them a mass equal to 0 and to normalizing the weights of the remaining correctly classified samples, in order to keep a probability measure. Formally,
\begin{equation}\label{eq:hard_weighting_strat}
    \forall i \in \llbracket 1, N \rrbracket, a_i = \left\{
    \begin{array}{ll}
        0 & \mbox{ if } \yy_t = f_\theta(\xx_i),\\
        1/n & \mbox{ otherwise,}
    \end{array}
\right.
\end{equation}
where $n$ is the number of samples such that $\yy_t \neq f_\theta(\xx)$. This strategy has the benefit that it can be used in the original OT cost without further modifications as we only change the probability weights of the empirical distributions. However, such a strategy would rely on the performances of the classifier to correctly detect all outliers and to correctly classify all clean samples. This is why we propose another soft reweighting strategy through the ground cost.
%that the classifier has classified all clean samples correctly.

Our \emph{soft reweighting strategy} consists in associating a larger transport cost to detected outliers and to let robust OT costs give them a smaller transport weight. To achieve this, we propose to modify the ground cost by considering the discrepancy between the classifier output of a source sample and the target label and vice-versa. %Denote the label of source samples as $\yy_s$ and the label of target samples as $\yy_t$.
We change the usual Euclidean distance as follows:

\begin{equation}\label{eq:new_ground_cost}
    c_\gamma^\prime(\xx_i, \zz_j) = \|\xx_i - \zz_j\|_2 + \frac{\gamma}{\operatorname{CE}(\yy_s, f_\theta(\zz_j))} + \frac{\gamma}{\operatorname{CE}(\yy_t, f_\theta(\xx_i))},
\end{equation}
where $\xx_i$ is a source sample, $\zz_j$ is a target sample and $\gamma$ a positive coefficient. We now give an intuition on how the new ground cost $c_\gamma^\prime(\cdot, \cdot)$ works. When the classification output $f_\theta(\zz_j)$ (resp. $f_\theta(\xx_i)$) is different from the label $\yy_s$ (resp. $\yy_t$), the cross-entropy loss is large and the new terms in the ground cost are small, which leads the cost to be close to the Euclidean distance. However, if the sample $\xx_i$ (resp. $\zz_j$) is a second type outlier that has been detected by the classifier, the prediction $f_\theta(\xx_i)$ (resp. $f_\theta(\zz_j)$) would be equal to $\yy_t$ (resp. $\yy_s$) and the extra terms would then be large, leading to a sample costly to transport. Thus, it would not be transported by a robust OT variant like first type outliers. In the case where some samples are close to the decision boundary, their associated cost would be bigger but they would still be partially transported with respect to $\tau$. Note that while our two strategies are related by reducing the probability weights of detected outliers, they are not entirely equivalent because of the renormalization step of the hard weighting strategy.

\paragraph{Computation of strategies}

Our hard weighting strategy changes the input measure weights without modifying the definition of optimal transport costs. That is why it can be used in any OT cost $h$ defined in Section \ref{sec:rw}, as well as their dual, to make them robust to second type outliers. When used in the robust OT variants, these variants are then robust to the two types of outliers, as we show in our experiments. Our soft weighting strategy can also be used in any primal OT variant defined in Section \ref{sec:rw}. However,  as our soft weighting strategy modifies the ground cost, the dual formulations of these OT costs change as well. That is why for many applications which rely on the dual formulation of optimal transport, we cannot use the soft weighting strategy directly and rely on the hard weighting strategy instead. We compare the different computation of our strategies in Algorithms \ref{alg:hard} and \ref{alg:soft}.

%\kf{TO do add algorithm (hard weighting and soft weighting in parallele}

%
%\vspace{-0.2cm}
%\begin{comment}
\begin{minipage}{0.46\textwidth}
\begin{algorithm}[H]
    \centering
    \caption{Hard weighting strategy}\label{alg:hard}
    \begin{algorithmic}[1]
        \State \text{Data processing}
        \State \text{Train classifier $f_\theta$ with $L_{\text{AR}}$ (Eq. (\ref{eq:AR}))}
        \State \text{Define $\boldsymbol{a}^\prime, \boldsymbol{b}^\prime$ as in Eq. (\ref{eq:hard_weighting_strat})}
        \State \text{Compute OT cost $h_c(\boldsymbol{a}^\prime, \boldsymbol{b}^\prime)$}
    \end{algorithmic}
\end{algorithm}
\end{minipage}
\hfill
\begin{minipage}{0.46\textwidth}
\begin{algorithm}[H]
    \centering
    \caption{Soft weighting strategy}\label{alg:soft}
    \begin{algorithmic}[1]
        \State \text{Data processing}
        \State \text{Train classifier $f_\theta$ with $L_{\text{AR}}$ (Eq. (\ref{eq:AR}))}
        \State \text{Compute $c^\prime$ as in Eq. (\ref{eq:new_ground_cost})}
        \State \text{Compute OT cost $h_{c^\prime}(\boldsymbol{a}, \boldsymbol{b})$}
    \end{algorithmic}
\end{algorithm}
\end{minipage}
%\vspace{-0.2cm}
%\end{comment}

\paragraph{2D example}
We illustrate our methods on a 2D example in Figure \ref{fig:toy_classif_outlier}. We consider the same example and setting as in Figure \ref{fig:toy_uot_outlier}. The classifier trained without adversarial regularization is not able to detect second type outliers. This leads our solution to be equivalent to the unbalanced OT cost and to transport these outliers. However, when we consider the adversarial regularization, we can see that these outliers are detected. Thus the hard weighting strategy remove their mass and their influence on the transport problem for all values of $\tau$. Regarding the soft reweighting strategy, for large values of $\tau$, we still transport some mass of outlier samples but when $\tau$ decreases, we only transport clean samples. This toy experiment shows the interests of our strategies to mitigate the influence of outliers.

\paragraph{Limitations}
The limitations of our work are that our methods highly depend on the ability of the classifier to detect outliers and to correctly classify clean samples. In the case where none of the outliers are detected and all clean samples are correctly classified, we recover the considered OT cost and we would be sensitive to the outliers of second type.  In the experimental section, we show such a case on one of our experiment, where our method based on the classifier trained with cross-entropy behaves like the Wasserstein distance. If the classifier is unable to classify clean samples, as for instance when $\alpha$ and $\beta$ measures overlap, we might then suffer even more from outliers as they would get more influence in the probability measures. Finally, as we base our work on a label noise robust learning strategy, it can be possible to use other state-of-the-art method \citep{Li2020DivideMix, NEURIPS2018_a19744e2} and we let such comparison as future work.

\begin{figure}[t]
  \begin{center}
    \includegraphics[width=1.\linewidth]{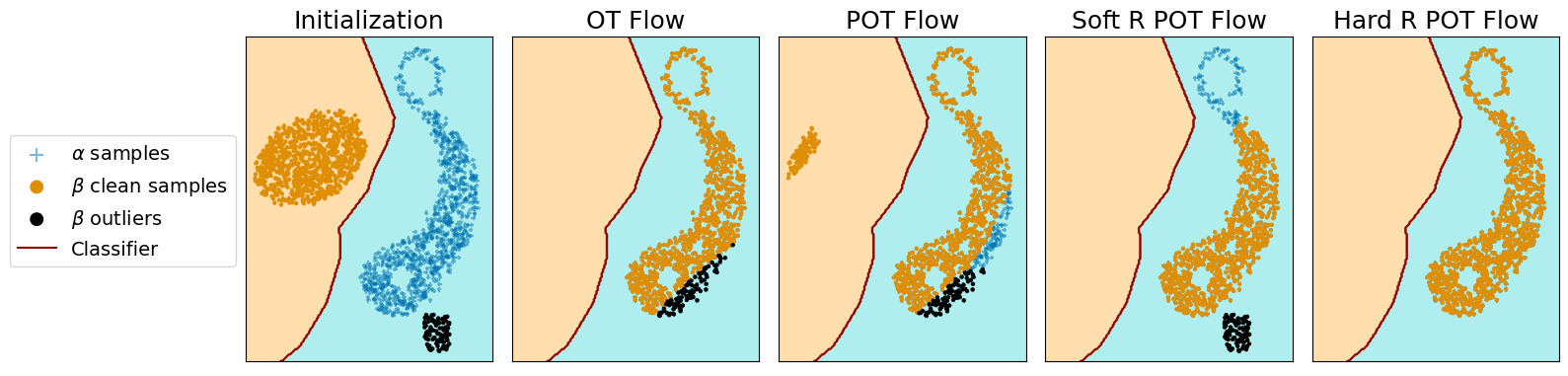}
  \end{center}
  \vspace{-0.cm}
    \caption{Gradient flow on a 2D corrupted dataset for different optimal transport cost. $\beta$ is corrupted with 10$\%$ of noisy samples (in black color) closer to the $\alpha$ measure than clean $\beta$ samples. We plot different converged flows which show that our methods do not transport outliers contrary to others.}
    \label{fig:gen_imgs_gradient_flow_2D}
  \vspace{-0.cm}
\end{figure}

\section{Experiments}\label{sec:experiments}
In this section, we evaluate our methods on several experiments. We call our method SROT, which stands for Strengthened Robust OT with soft or hard weighting (w.). We show that SROT results in cleaner gradient flows, correctly generates images in the presence of noise, and improves label propagation as compared to previous robust OT variants. We used the Pytorch library \citep{paszke2017automatic} to design our experiments and ran them on a single 1080-Ti NVIDIA GPU. Regarding the computation of OT, we used the Python OT package \citep{Flamary2021}. Code will be released upon publication. Hyperparameters were validated on GAN experiments and used as is for other experiments except for the toy experiments, we give more details in the supplementary materials. %We call our method SROT which stands for Strengthened Robust OT with soft or hard weighting (w.).

\subsection{Gradient flow on synthetic data and CelebA dataset}
We start the experimental section with gradient flow experiments. Let $\alpha$ be a given measure, the purpose of gradient flows is to model a measure $\beta_t$ which at each iteration follows the gradient direction minimizing the loss $\beta_t \mapsto h(\alpha, \beta_t)$ \citep{COT_Peyre, liutkus19a}, where $h$ stands for any OT cost (W, $\operatorname{OT}_{\texttt{TV}}^{\tau, \varepsilon}, \operatorname{ROT}_\rho$). The gradient flow can be seen as a non parametric data fitting problem where the modeled measure $\beta$ is parametrized by a vector position $\xx$ that encodes its support. We follow the same experimental procedure as in \cite{feydy19a}. The gradient flow algorithm is based on an explicit Euler integration scheme which integrates at each iteration an ODE from some initial measure at $t=0$, %As the OT costs take are uniform distributions as inputs, we have an inherent bias when we calculate their gradient that we correct by multiplying it by $n$.
\emph{i.e.,} $\dot{\ZZ}(t)=-n \nabla_{\boldsymbol{\ZZ}}h(\frac{1}{n}\sum_{i} \delta_{\xx_i}, \frac{1}{n}\sum_{j} \delta_{\zz_j(t)})$.
We consider gradient flows on a 2D dataset and on female images from CelebA dataset using the following OT losses: the Wasserstein distance, partial OT \citep{chapel2020partial} which is equivalent to  the robust OT cost from \cite{mukherjee2020outlierrobust}  (see their third formulation) and our strategies. SROT hard weighting using the Wasserstein distance and SROT soft relies on Partial OT.

\paragraph{2D synthetic dataset}
We show the gradient flow of the different OT costs in Figure \ref{fig:gen_imgs_gradient_flow_2D}. We generate 1000 2D samples for $\alpha$ and $\beta_0$. 10 $\%$ of $\beta$ samples (in black color) are in fact second type outliers closer to $\alpha$ than clean samples from $\beta_0$. We used a learning rate of 0.01 and 400 iterations. For Partial OT, we set the mass to transfer to $90\%$ as it is the quantity of clean samples.

The results of the different gradient flows show that the Wasserstein distance and the robust OT variants transport the outlier samples. As expected partial OT even performs worse as some clean samples are not transported while outliers are. However our two reweighting strategies are robust against outliers because our considered classifier detects all of them as shown in Figure \ref{fig:gen_imgs_gradient_flow_2D}. By its renormalization step, our hard weighting strategy, which relies on the Wasserstein distance, is even able to match perfectly the shape of the $\alpha$ measure. The soft strategy however only partially recovers it because it relies on partial OT and only transports $90\%$ of the measures. Furthermore, we see that the outliers are the most expensive samples to transport. The different results between our methods show that our formulations are not equivalent. In the next paragraph, we show how a simple rescaling of measures can lead to equivalent results between our strategies and detail it in supplementary.

\begin{figure}[t]
  \begin{center}
    \includegraphics[width=1.\linewidth]{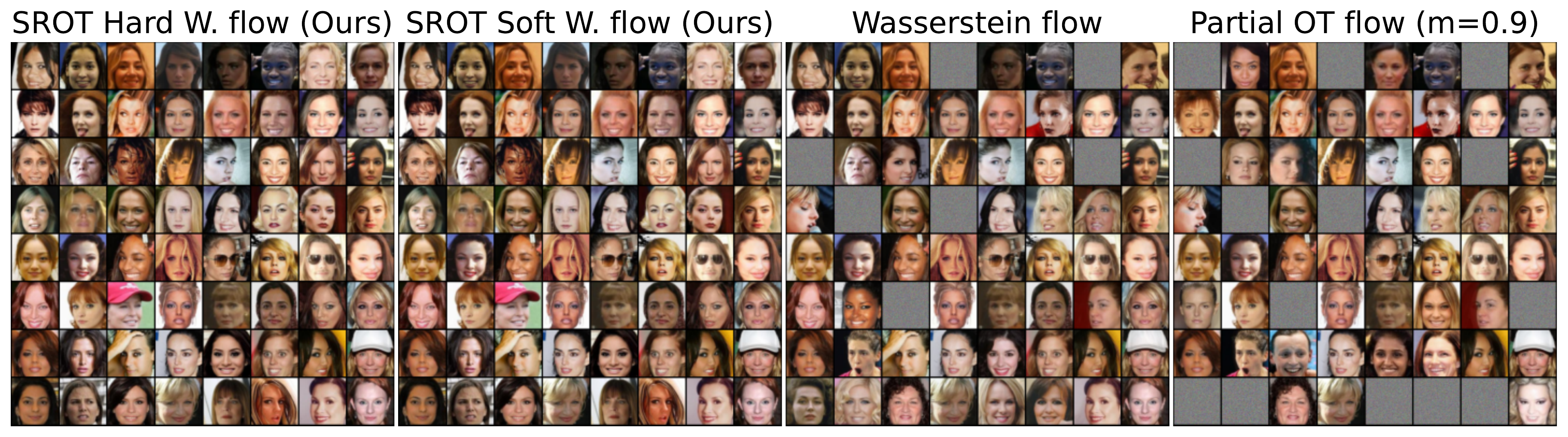}
  \end{center}
  \vspace{-0.25cm}
    \caption{Converged gradient flow on the CelebA dataset for different OT losses. $\beta_0$ samples are 3000 random samples while $\alpha$ samples are 2700 clean female images and 300 random samples. Our SROT methods do not converged to any Gaussian random samples contrary to other OT flows.}
    \label{fig:gen_imgs_gradient_flow}
  \vspace{-0.3cm}
\end{figure}

%Check jumbot
%\kf{Discuss Problem}

%\kf{Discuss Setting + competitor}

%\kf{Discuss results}

\paragraph{High dimensional dataset}
%\kf{Discuss Setting + competitor}
Similarly to \citep{liutkus19a}, we consider the CelebA dataset \citep{liu2015faceattributes}, which is a face attributes dataset composed of 202,599 images. Our measure $\alpha$ is composed of 2700 female images, corresponding to the clean measure $\alpha_c$, and 300 Gaussian random samples as outliers while the $\beta$ measure is composed of 3000 Gaussian random samples. On this dataset, our considered classifier trained with adversarial regularization (AR) correctly classifies clean samples and detect all outliers while the classifier trained with cross-entopy only correctly classifies clean samples but fails to detect outliers. We provide in the supplementary materials the different ground costs of our soft reweighting methods which show that outliers are indeed detected as they have the largest costs. We perform the gradient flow of the same OT costs as above but we consider a bigger variety of transferred mass for partial OT. We also consider SROT based on a classifier trained only with CE (SROT-CE) and we show that it does not perform as good as SROT with AR classifier. We rescale the mass of $\alpha$ to 0.9 before making the SROT soft weighting flow to show that the results of our methods are equivalent and we provide results without this rescaling in supplementary.

\begin{figure}[t]
  \begin{center}
    \includegraphics[width=1.\linewidth]{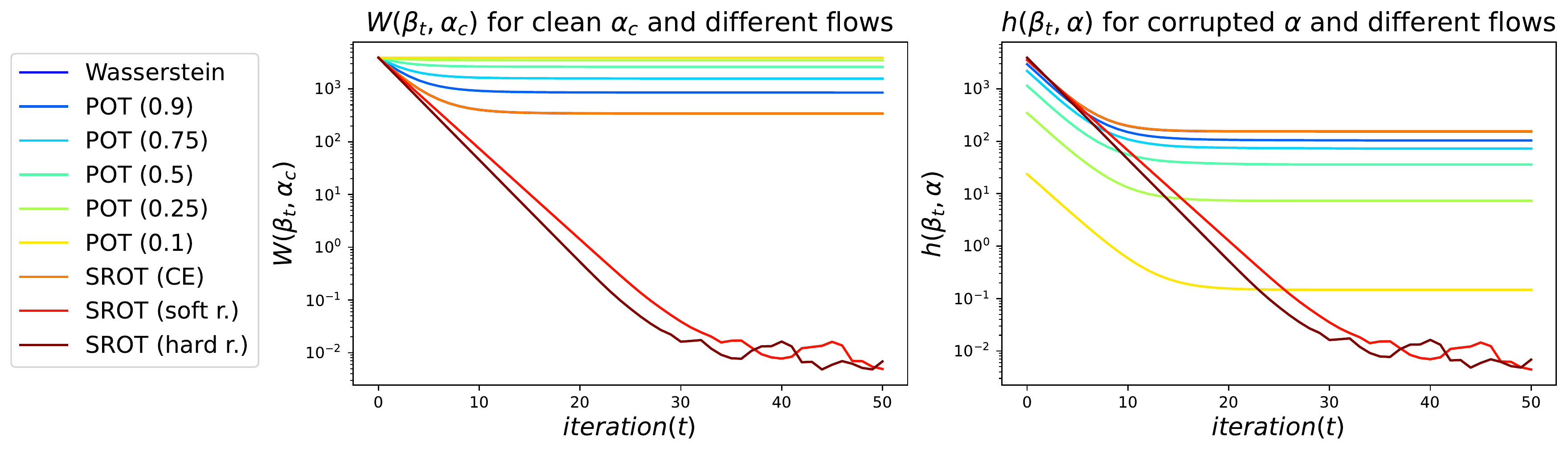}
  \end{center}
  \vspace{-0cm}
    \caption{Wasserstein distance between gradient flow measures from different optimal transport costs and clean $\alpha_c$ ($W(\alpha_c, \beta_{t})$) and as well as loss value ($h(\alpha, \beta_{t})$) on the CelebA dataset.  $\beta_0$ samples are 3000 random samples while $\alpha$ samples are 2700 clean female images and 300 random samples. SROT (soft and hard) are able to recover $\alpha_c$ contrary to other flows while they minimize their loss.}
    \label{fig:gradient_flow_wass_values}
  \vspace{-0cm}
\end{figure}

We evaluate the results of the flows qualitatively in Figure \ref{fig:gen_imgs_gradient_flow}. We can see that the final measures of our two methods $\beta_{\text{final}}$ do not have any Gaussian random samples contrary to the Wasserstein flow and the partial OT flow. To quantitatively compare how the different flows handled outliers, we plot the Wasserstein distance between each flow and the clean $\alpha_c$ measure, \emph{i.e.,} $W(\alpha_c, \beta_{t})$ (right plot of Figure \ref{fig:gradient_flow_wass_values}). We also plot the value of the different losses, \emph{i.e.,} $h(\alpha, \beta_{t})$, to check that each flow minimizes its loss (left plot of Figure \ref{fig:gradient_flow_wass_values}). While each flow minimizes its loss function, only the flow of our methods are able to fit the clean distribution $\alpha_c$, showing that they were able to mitigate the influence of outliers contrary to other flows. The difference of speed of convergence between our methods is due to the rescaled measures $\alpha$ for our soft weighting strategy. Finally, we note that SROT-CE performs similarly to the Wasserstein flow because the classifier does not detect outliers.

\subsection{Monge map experiment}\label{app_sec:monge_map}

\begin{figure}[t]
  \begin{center}
    \includegraphics[width=1.\linewidth]{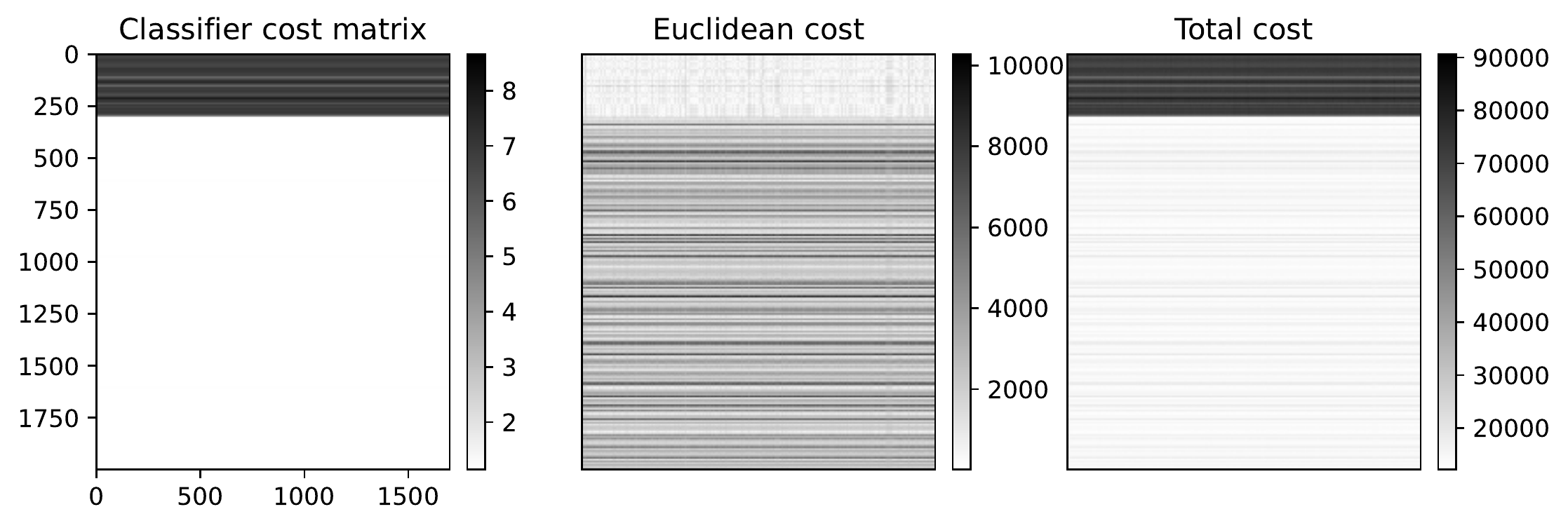}
  \end{center}
  \vspace{-0cm}
    \caption{Different ground costs which composed the full ground cost used in our SROT soft weighting method. The first 300 source samples are outliers. (Left plot) is the ground cost representing the new terms based on the classifier prediction. (Middle plot) represents the Euclidean distance between all source and target samples. (Right plot) is our total ground cost with $\gamma = \operatorname{max_{1\leq i,j \leq n}}(\|\xx_i - \yy_j\|_{2})$.}
    \label{fig:all_ground_costs_MS}
  \vspace{-0cm}
\end{figure}

In this section, we look for an optimal transport map which transforms a measure into another following \cite{seguy2018large}. To get such a map, we can rely on the optimal transport plan of the Kantorovich problem $\Pi$, which defines a map between the source and target samples through the computation of the so-called barycentric projection. The barycentric projection transports source (resp. target) samples to the target (resp. source) samples using the OT plan. For each target sample $\zz_j$, the corresponding source element can be computed as follows: $\Pi_{\zz_j} =\sum_{i=1}^n \pi_{i,j} \xx_i$. We could then define a map $T$ which maps a sample to its corresponding barycentric projection. In the case of second type outliers, it is likely that some samples $\zz_j$ would be matched to second type outliers present in the source dataset and we want to study the relevance of our SROT methods in this context. We give a longer introduction on Monge map approximation in the supplementary material.

\paragraph{Datasets}

We consider the class of digits "1" from the SVHN datasets \citep{svhn} as source domain and from the MNIST dataset as target domain. The source dataset is composed of 1700 clean "1" SVHN digit samples and 300 outlier samples corresponding to "1" MNIST digits. The target dataset is composed of 1700 "1" MNIST digits. The noise rate is set to $15\%$.

\begin{figure}[t]
  \begin{center}
    \includegraphics[width=1.\linewidth]{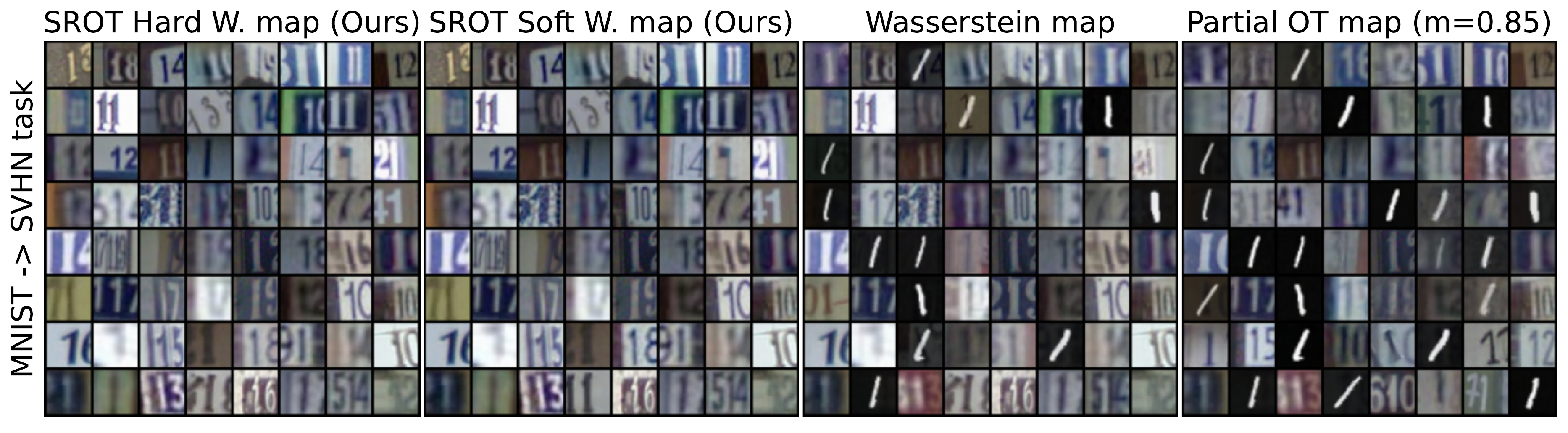}
  \end{center}
  \vspace{-0.cm}
    \caption{Converged Monge map $T_{\theta^\star}$ on the MNIST to SVHN task for different OT losses. $\alpha$ samples are 1700 clean SVHN images and 300 MNIST second type outliers while $\beta$ samples are 1700 MNIST samples. Our classifier detects MNIST samples present in the source dataset. Our SROT methods do not generate any MNIST samples contrary to other OT map.}
    \label{fig:gen_imgs_gradient_flow_MS}
  \vspace{-0.cm}
\end{figure}

\begin{figure}[t]
  \begin{center}
    \includegraphics[width=1.\linewidth]{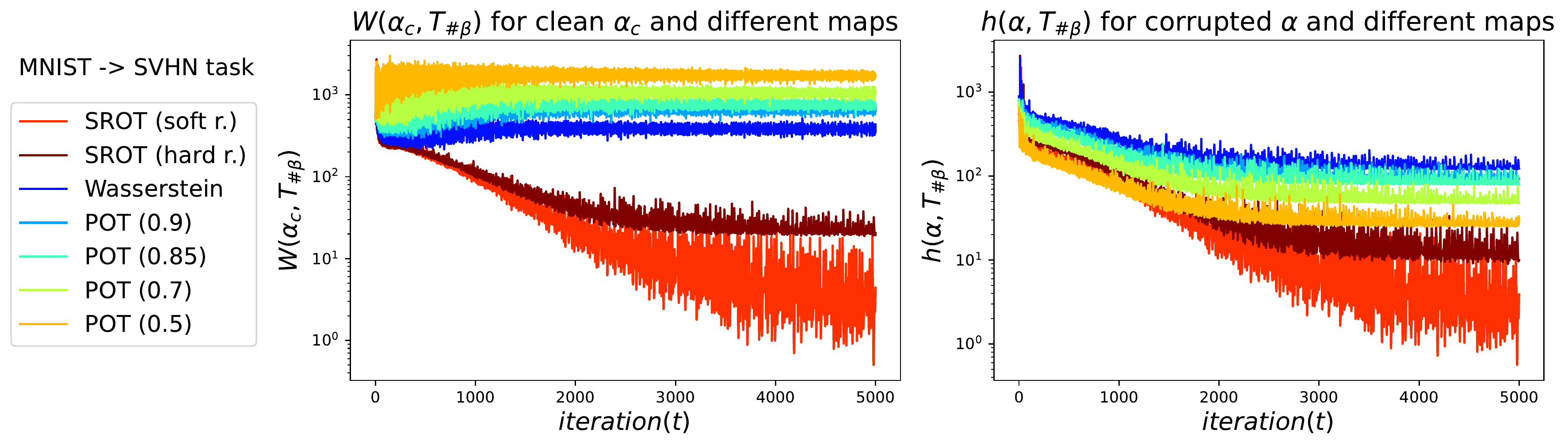}
  \end{center}
  \vspace{-0.cm}
    \caption{Wasserstein distance between push-forward measure $T_{\#\beta}$ from different optimal transport costs and clean $\alpha_c$ ($W(\alpha_c, T_{\#\beta})$) and as well as loss value ($h(\alpha, T_{\#\beta})$) on MNIST$\rightarrow$SVHN.  $\beta$ samples are 1700 MNIST samples while $\alpha$ samples are 1700 SVHN images and 300 MNIST samples. Our methods SROT hard and soft weighting are able to match the clean source distribution $\alpha_c$.}
    \label{fig:gradient_flow_wass_values_MS}
  \vspace{-0.cm}
\end{figure}

\paragraph{Results}

The purpose of our experiment is to transform the target samples in clean source samples and to not transform target MNIST samples into source outlier MNIST samples. To achieve this, we use the optimal transport plan from several OT variants. We compare our SROT methods with the Wasserstein distance and the Partial OT cost (amount of transported mass $m=\{0.9, 0.85, 0.7, 0.5\}$). Our SROT hard weighting method uses the Wasserstein distance while the SROT soft weighting method uses the Partial OT cost with $m=0.85$. Note that we use our rescaling trick for our SROT soft weighting method.

The classifier that we consider to classify source and target samples is able to detect the second type MNIST outliers. To demonstrate this, we show the different ground costs associated with SROT soft weighting in Figure \ref{fig:all_ground_costs_MS}. The ground costs associated to the classifier can be found in the left plot of these figures. One can see that the 300 first source samples have a larger cost than the remaining samples and these 300 samples are the outliers. It shows that we are able to detect the outliers. Furthermore, the Euclidean distance between outliers and target samples is smaller than the Euclidean cost between target and clean source samples,  showing that outliers are of second type. Now that we have detected outliers, we want to check if we do not match any of them.

We can qualitatively assess our methods by looking at the produced images of the converged map $T_{\theta^\star}$. The results for the MNIST to SVHN task can be found in Figure \ref{fig:gen_imgs_gradient_flow_MS}. We can see that contrary to the other methods, SROT hard and soft weighting do not generate any MNIST images. Note that some MNIST images do not have a background as black as it could be expected in the two figures because of the renormalizations between all samples when plotting the images.  %\kf{Results : Qualitative (images), quantitative, Wasserstein distance $W(T_{\#\alpha}, \bbeta)$.}

Then, we can have a quantitative metric similar to the one developed in the gradient flow experiment. Instead of computing the Wasserstein distance between the flow and the clean measure, we can compute the Wasserstein distance between the pushforward measure $T_{\theta \# \beta}$ and the clean measure $\alpha_c$. Indeed, if we are able to detect outliers and remove their influence from the problem, we should be able to match the clean measure. The results can be found in Figures \ref{fig:gradient_flow_wass_values_MS}. We also plot the loss value to show that the loss is minimized for each method.

Results show that the only methods which get very close to the clean source measures $\alpha_c$ are our SROT methods. This shows that our methods largely mitigate the impact of outliers compare to competitors. We also note that in this case the soft weighting strategy performs slightly better than the hard weighting strategy. It is because the classifier used in the soft strategy detected all outliers while the one used in the hard weighting strategy detected 0.999$\%$ of them. Thus the hard weighting strategy transported 0.001$\%$ of the mass of all outliers. Finally, we empirically saw that all competitor methods fully transported the $15\%$ of outliers contrary to our methods.

\begin{table}[t!]
\small
\begin{center}
\begin{tabular}{ |c|c|c|c|c|c|c| }
 \hline
 Noise rate & SROT (H-AR) & WGAN-GP & ROT (0.1) & ROT(0.25) & ROT(0.5) & ROT(0.75) \\
 \hline
  %5$\%$ & \textbf{4.80 $\pm$ 0.2} & 3.71 $\pm$ 0.09 & 4.07 $\pm$ 0.1 & 3.48 $\pm$ 0.1 & 3.63 $\pm$ 0.1 & 3.50 $\pm$ 0.04\\
 %10$\%$ & \textbf{5.0 $\pm$ 0.1} & 3.56 $\pm$ 0.06 & 3.82 $\pm$ 0.1 & 3.85 $\pm$ 0.16 & 3.85 $\pm$ 0.08 & 3.30 $\pm$ 0.09\\
 %15$\%$ & \textbf{4.85 $\pm$ 0.1} & 3.08 $\pm$ 0.04 & 3.10 $\pm$ 0.07 & 3.07 $\pm$ 0.09 & 3.32 $\pm$ 0.09 & 2.54 $\pm$ 0.05\\
   5$\%$ & \textbf{4.80 $\pm$ 0.2} & 3.98 $\pm$ 0.16 & 3.93 $\pm$ 0.08 & 3.64 $\pm$ 0.08 & 4.07 $\pm$ 0.1 & 3.52 $\pm$ 0.10\\
 10$\%$ & \textbf{4.95 $\pm$ 0.2} & 3.48 $\pm$ 0.11 & 3.92 $\pm$ 0.09 & 3.38 $\pm$ 0.08 & 3.67 $\pm$ 0.08 & 2.96 $\pm$ 0.09\\
 15$\%$ & \textbf{4.99 $\pm$ 0.2} & 2.89 $\pm$ 0.06 & 2.61 $\pm$ 0.10 & 3.36 $\pm$ 0.11 & 2.36 $\pm$ 0.07 & 2.58 $\pm$ 0.06\\
 \hline
\end{tabular}
\end{center}
\caption{(Best scores in \textbf{bold}, higher is better) Inception scores from different GANs trained on corrupted Cifar10 dataset. The dataset is corrupted with different percentage of noisy samples.}
\label{tab:inception_scores}
\vspace{-0.cm}
\end{table}

\begin{figure}[t]
  \begin{center}
    \includegraphics[width=1.\linewidth]{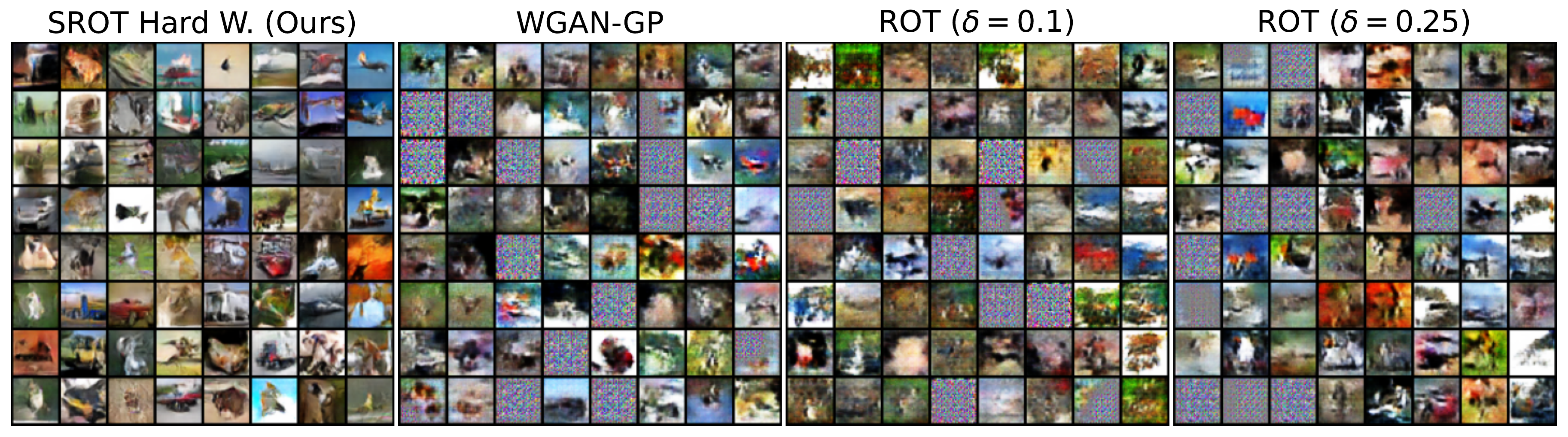}
  \end{center}
  \vspace{-0cm}
    \caption{Generated images from different GANs trained on a corrupted Cifar10 dataset with $15\%$ of noisy samples. SROT does not generate any random images as opposed to other GANs.}
    \label{fig:gan_generated_imgs}
  \vspace{-0.cm}
\end{figure}

\subsection{Generative Adversarial Networks on Cifar10}\label{sec:exp_gan}
We consider now the problem of Generative Adversarial Networks \citep{goodfellow_gan} where one wants to generate images which look like real images. The problem takes the form of a data fitting problem where we want to fit a parametric model $\beta_\theta$, \emph{i.e.,} images generated from a generator, to some empirical measure $\alpha$, \emph{i.e.,} the measure of real images. We consider the case where $\alpha$ is tainted with samples relatively similar to $\beta_\theta$ (see supplementary for more details). The different noise rate are 5$\%$, 10$\%$ and 15$\%$. Our goal is to find the best $\theta$ which minimizes $W(\alpha, \beta_\theta)$.%, \emph{i.e.,} $\text{min}_\theta W(\alpha, \beta_\theta)$.
%We artificially create these samples by generating samples from the generator that we perturbed using a normal distribution $\mathcal{N}(0, \sigma^2)$ where $\sigma^2$ is generated using a signal-to-noise ratio, $\sigma = \frac{\|X\|^2}{\|\epsilon\|^2}$ with $X\sim \beta_\theta$. We denote this distribution $\alpha_o$, the clean distribution $\alpha_c$ and then we consider the distribution $\alpha = (1-\kappa) \alpha_c + \kappa \alpha_o$, where $\kappa$ is the noise rate that we consider of 5$\%$, 10$\%$ and 15$\%$.

We consider the dataset of images Cifar10 \citep{Cifar10}. We associate our method with the general robust optimal transport variant of \cite{RobustOptimalTransport2022Nietert} ($\delta=0.1$) which generalizes the work of \cite{balaji2020robust}. It can be computed with an easy dual formulation close to the Kantorovich-Rubinstein duality used in WGAN \citep{arjovsky17a, Gulrajani2017}. As we rely on a dual formulation, we only use our hard weighting strategy as it does not require further modification. We compare our method to the robust OT GAN variant \cite{RobustOptimalTransport2022Nietert} as well as WGAN  and use the inception score as quantitative metric. Regarding training procedure and architecture of the generator and critic, we considered the setting developed in \cite{NIPS2017_dfd7468a}.

In this setting, we train our classifier on a dataset composed of 50,000 images generated from our generator as source measure and the corrupted Cifar10 dataset as target measure. Our classifier trained with adversarial training is able to detect all outliers present in the dataset while it is not the case for the classifier trained with only CE, thus SROT-CE is equal to ROT. We compute the inception scores of the different methods  and gather results in Table \ref{tab:inception_scores}. We see that our method outperforms other generative models by more than 1$\%$ in all noise setting (1.5 $\%$ in the 15$\%$ setting). We then evaluate the performances of the different methods qualitatively by generating samples from different GANs (see Figure \ref{fig:gan_generated_imgs}). We see that our methods do not generate any noisy samples while WGAN-GP and ROT produces several noisy samples as well as samples which are both noisy and realistic.

\subsection{Label propagation on digits datasets}

\begin{minipage}[t]{0.3\textwidth}
  \centering\raisebox{\dimexpr \topskip-\height}{%
    \includegraphics[width=1.\linewidth]{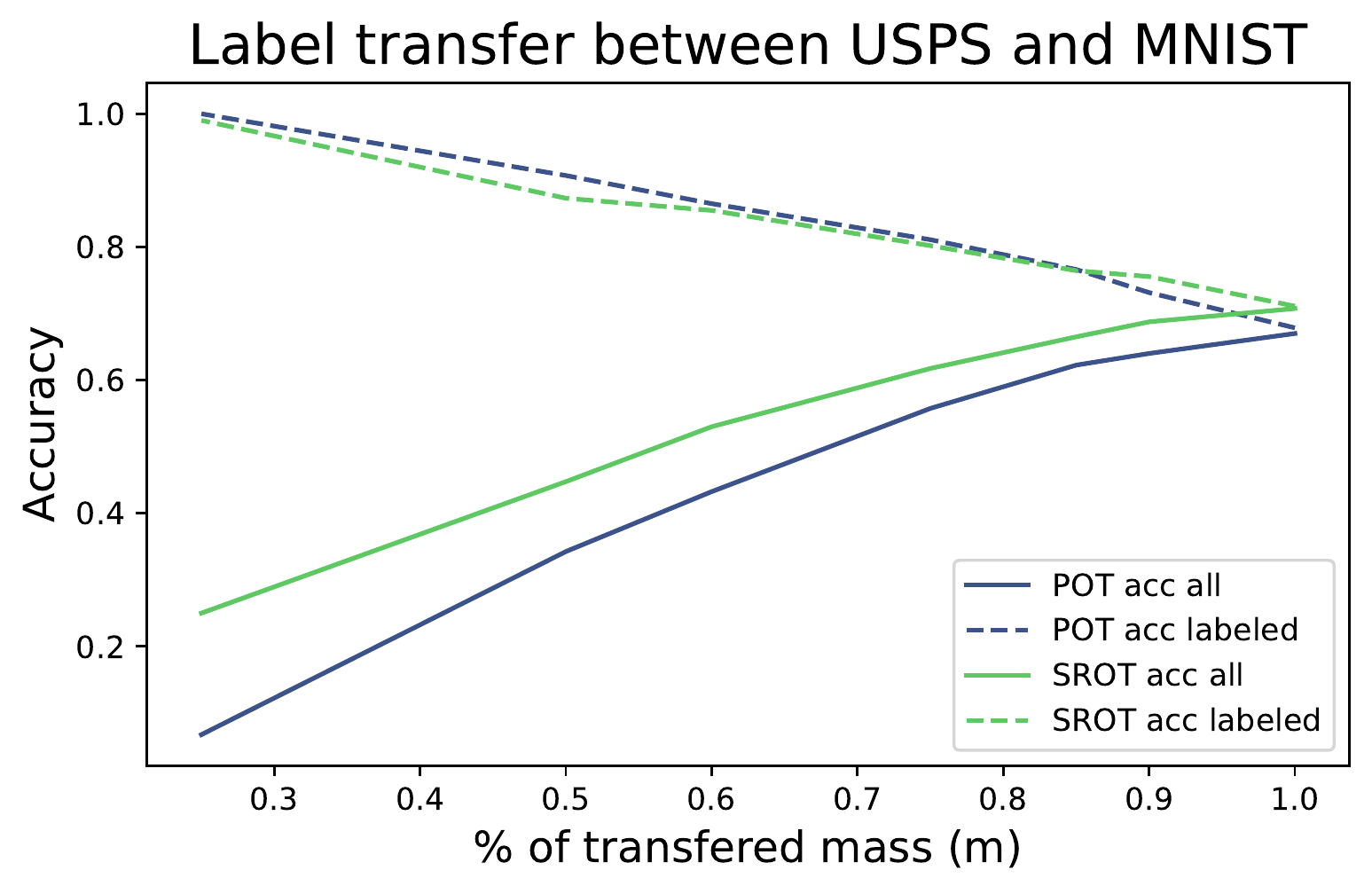}}
  \captionof{figure}{Accuracy on the clean target domain for several OT variants. SROT outperforms ROT by at least 2$\%$.}
  \label{fig:acc_label_transfer}
\end{minipage}\hfill
\begin{minipage}[t]{0.67\textwidth}
  We consider a classification problem on corrupted digits dataset following the experiments from \citep{mukherjee2020outlierrobust, chapel2021unbalanced}. Let the source dataset be composed of 400 USPS \citep{USPS} samples from each class 0 and 1 (200 per class) and 100 samples from class 2 of the Fashion MNIST dataset \citep{fashionmnist}. The target measure is composed of 200 samples from each class 0 and 1 of MNIST \cite{MNIST} and 100 samples from each class 0 and 1 of the USPS dataset (50 per class). We compute the OT plan between datasets and we classify a target sample by propagating the label of the source sample it is the most transported to, as long as the transported mass is at least equal to 0.25$b_j$ similarly to \cite{chapel2021unbalanced}.
\end{minipage}

We compare our hard reweighting strategy based on Partial OT with the robust OT variant from \cite{mukherjee2020outlierrobust}. The accuracy is defined as the number of \emph{clean target samples} that are correctly classified divided by the total number of samples (400), and the labeled accuracy is the proportion of correctly classified samples among the classified samples. The classifier $f_\theta$ trained with adversarial regularization to classify source and target samples, is able to detect the USPS samples which are present in the target dataset. Our results are gathered in Figure \ref{fig:acc_label_transfer} and they show that our method outperforms the other baseline by more than 2$\%$ on average. Furthermore as the percentage of transported mass increases, the overall accuracy increases as more samples are correctly classified while the labeled accuracy decreases as more outliers are classified.

%\kf{Discuss Problem}
%\kf{Discuss Setting}
%\kf{Discuss results}

\section{Conclusion and broader impact}\label{sec:conclusion}
Outliers are typically defined as samples costly to move in the OT literature. In this paper, we showed that it is an incomplete definition as it does not consider samples close to the target measure. The robust OT variants fail to tackle these outliers as they keep transporting them. To make these variants robust to this second type of outliers, we proposed to rely on a classifier to predict if a sample is more likely to belong to the source or the target measure. In this context, the second type of outliers can be seen as noisy label samples, and we trained the classifier with adversarial training to make them robust against them. After detecting these outliers with the classifier, we proposed to mitigate their influence in the transport problem by using new empirical distributions or a modified ground cost. We finally demonstrated empirically that we can detect these outliers and remove their influence from the transport problem. To the best of our knowledge, this work does not lead to any negative outcomes either in ethical or societal aspects, as the contributions of this paper are semantic on the notion of outliers and methodological to detect and alleviate their influence in the OT problem.

\begin{ack}
This work was partially supported by NSERC Discovery grant (RGPIN-2019-06512), a Samsung grant and a Canada CIFAR AI chair. KF thanks Quentin Bertrand, Charles Guille-Escuret, Thibault S\'ejourn\'e, Tiago Salvador, Alex Tong, Hiroki Naganuma and Ryan D'Orazio for fruitful feedbacks and discussions on the early manuscript. KF also thanks his partner for her support.
\end{ack}

\newpage

\bibliography{collas2022_conference}
\bibliographystyle{collas2022_conference}

\newpage

\appendix

\paragraph{\Huge{Supplementary material}}

\paragraph{Outline.} The supplementary material of this paper discusses the connections between our strategies, furnishes new experiments (\textbf{gradient flow} and \textbf{Monge map estimation}) and all the experimental procedures. It is organized as follows:
\begin{itemize}
    \item In Section \ref{app_sec:relationship}, we first review the connections between our strategies and furnish additional experiments on gradient flows.
    \item In Section \ref{app_sec:exp_details}, we give the implementation details of all experiments present in the main paper except Monge map.
    \item In Section \ref{app_sec:monge_map}, we give a longer introduction, furnish new experiments and experimental procedures on Monge map estimation between the MNIST dataset and the MNIST-M dataset.
\end{itemize}

\section{Discussion on SROT formulations}\label{app_sec:relationship}
In this section, we first discuss how we select $\gamma$ in order to ensure that outliers have a larger moving cost than clean samples. We then discuss in more details the relationship between our two formulations and show how they can achieve the same results using a rescaling of measures for the soft variants. Finally, we show the gradient flow of our soft weighting method without the rescaling of measures.
\paragraph{SROT soft weighting costs}

\begin{figure}[t]
  \begin{center}
    \includegraphics[width=1.\linewidth]{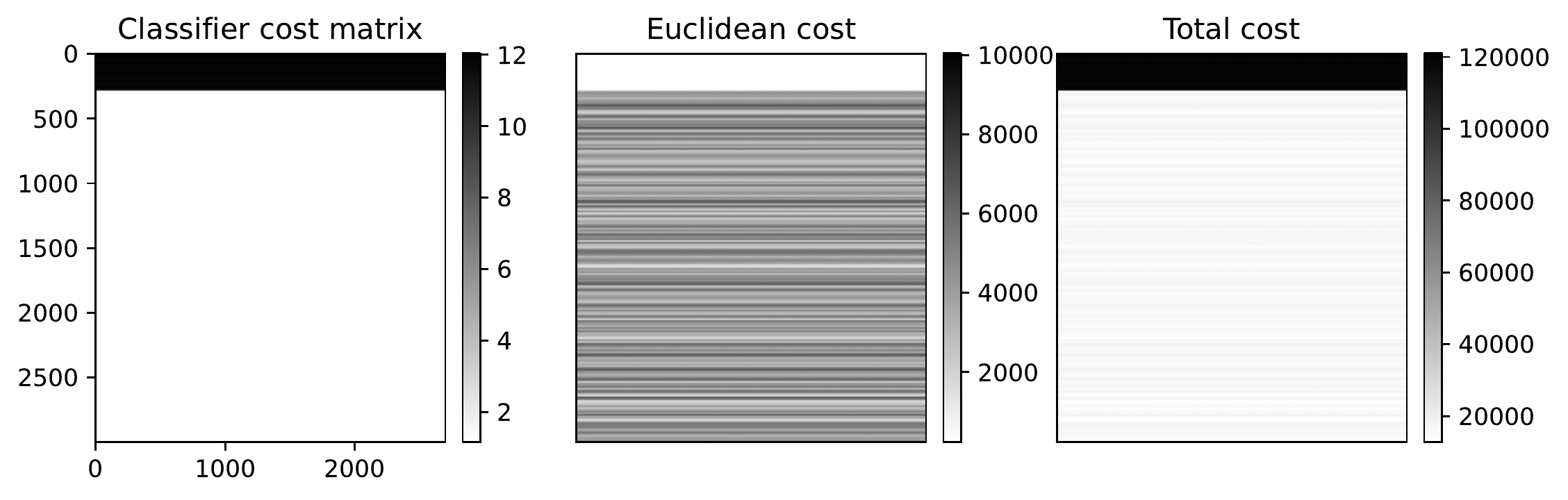}
  \end{center}
  \vspace{-0.cm}
    \caption{\textbf{(Gradient flow on CelebA)} Different ground costs which composed the ground cost used in our SROT soft weighting method. The first 300 target samples are outliers. (Left plot) is the ground cost representing the new terms based on the classifier prediction. (Middle plot) represents the Euclidean distance between all source and target samples. (Right plot) is our total ground cost with $\gamma = \operatorname{max_{1\leq i,j \leq n}}(\|\xx_i - \yy_j\|_{2})$.}
    \label{fig_app:all_ground_costs}
  \vspace{-0cm}
\end{figure}

Our soft weighting strategy relies on the principle that we are able to detect and increase the moving cost of outliers in order to not transport them. This is why we introduced our new ground cost:
\begin{equation}\label{app_eq:new_ground_cost}
    c_\gamma^\prime(\xx_i, \zz_j) = \|\xx_i - \zz_j\|_2 + \frac{\gamma}{\operatorname{CE}(\yy_s, f_\theta(\zz_j))} + \frac{\gamma}{\operatorname{CE}(\yy_t, f_\theta(\xx_i))},
\end{equation}
where $\yy_s$ (resp. $\yy_t$) is the source (resp. target) sample label. We study the relevance of this new ground cost with the gradient flow experiment on CelebA dataset. The two new last terms, $\frac{\gamma}{\operatorname{CE}(\yy_s, f_\theta(\zz_j))} + \frac{\gamma}{\operatorname{CE}(\yy_t, f_\theta(\xx_i))}$, would be large if we are able to detect outliers with the classifier and we plot their value in the left plot of Figure \ref{fig_app:all_ground_costs}. We can see that the 300 first source samples, which are the outliers, have the largest moving cost. Furthermore, as shown in the middle plot of Figure \ref{fig_app:all_ground_costs}, the Euclidean cost between $\beta$ samples and these outliers is a thousand time smaller than the Euclidean cost between $\beta$ samples and clean $\alpha_c$ samples. It shows that outliers are second type outliers. To ensure that outliers have a larger cost than clean samples, we can set $\gamma = \operatorname{max_{1\leq i,j \leq n}}(\|\xx_i - \yy_j\|_{2})$ and then outliers would have larger costs than clean samples as shown in the right plot of Figure \ref{fig_app:all_ground_costs}. In the next paragraph, we discuss the connections between our two formulations.

\paragraph{Equivalence of results} In this section, we discuss the equivalence between the two weighting strategies of our method SROT. We recall the different computations of our algorithms:

\begin{minipage}{0.46\textwidth}
\begin{algorithm}[H]
    \centering
    \caption{Hard weighting strategy}\label{alg_app:hard}
    \begin{algorithmic}[1]
        \State \text{Data processing}
        \State \text{Train classifier $f_\theta$ with $L_{\text{AR}}$}
        \State \text{Define $\boldsymbol{a}^\prime, \boldsymbol{b}^\prime$}
        \State \text{Compute OT cost $h_c(\boldsymbol{a}^\prime, \boldsymbol{b}^\prime)$}
    \end{algorithmic}
\end{algorithm}
\end{minipage}
\hfill
\begin{minipage}{0.46\textwidth}
\begin{algorithm}[H]
    \centering
    \caption{Soft weighting strategy}\label{alg_app:soft}
    \begin{algorithmic}[1]
        \State \text{Data processing}
        \State \text{Train classifier $f_\theta$ with $L_{\text{AR}}$ }
        \State \text{Compute $c^\prime$ }
        \State \text{Compute OT cost $h_{c^\prime}(\boldsymbol{a}, \boldsymbol{b})$}
    \end{algorithmic}
\end{algorithm}
\end{minipage}

The main difference between the two strategies is that the hard weighting strategy, with the Wasserstein distance, removes outliers from the empirical distributions, while the soft weighting strategy increases their moving cost and use a robust OT variant to not transport them. As such, the OT cost $h$ in the strategies is different and the input probability distributions are not equal, explaining why the two formulations are generally different. To explain similar results in the gradient flow experiment on the CelebA dataset, we illustrate that the SROT soft formulation based on Partial OT, with a rescaling of measures, can be seen as a Wasserstein flow between rescaled measures.

%The classifier is used in the two strategies to detect the proportion of outliers in datasets. For the gradient flow in high dimension where the distribution $\alpha$ has $10\%$ of outliers.
We explain how the different strategies work on the gradient flow experiment. The classifier trained with adversarial regularization was able to detect the $10\%$ outliers of the distribution $\alpha$. Thus what we want to do is to transport the full $\beta$ distribution to the remaining 90$\%$ of $\alpha$ samples corresponding to $\alpha_c$, \emph{i.e.,} $\alpha = \frac{1}{n} \sum_{i=1}^n \delta_{\xx_i}$ and $\beta = \frac{1}{N} \sum_{j=1}^N \delta_{\zz_j}$. The hard weighting strategy naturally removes these outliers and only keeps the 90$\%$ of clean samples, then it normalizes the measure to have a probability measure of clean samples. The soft weighting strategy however increases the moving cost of the outliers. We then rely on Partial OT to only transport clean samples with an amount of mass to transfer equal to $0.9$. However as the measures have different mass the measure $\beta$ ($\|\beta\|_1 = 1.$) would not be perfectly matched to the clean measure $\alpha_c$ ($\|\alpha_c\|_1 = 0.9$) at convergence.

One easy way to fix this is to rescale the $\beta$ mass to 0.9, \emph{i.e.,} $\|\beta\|_1 = 0.9$, as we did in our experiment. Then the empirical measures $\alpha_c$ and $\beta$ would have the same number of atoms and their atom the same weight, \emph{i.e.,} $\alpha_c = \frac{0.9}{N} \sum_{i=1}^N \delta_{\xx_i}$ and $\beta = \frac{0.9}{N} \sum_{j=1}^N \delta_{\zz_j}$. Using Partial OT we would entirely transport the rescaled $\beta$ to the rescaled $\alpha_c$, which is equivalent to a Wasserstein distance flow between rescaled $\beta$ and $\alpha_c$ measures. That is why we have similar results between our two strategies. The difference in the mass also explains the difference of convergence speed as the norm of gradients are smaller for the soft weighting strategy (see Figure 5 of the main paper). In the next paragraph, we give results of our SROT soft weighting strategy without this rescaling to show the differences in the flows.

\paragraph{SROT soft weighting results without rescaling}
In this experiment we show that without the measure rescaling, even if the classifier detects outliers, our SROT soft weighting strategy is not able to match the measure $\alpha_c$. This is similar to our 2D gradient flow example.

\begin{figure}[t]
  \begin{center}
    \includegraphics[width=1.\linewidth]{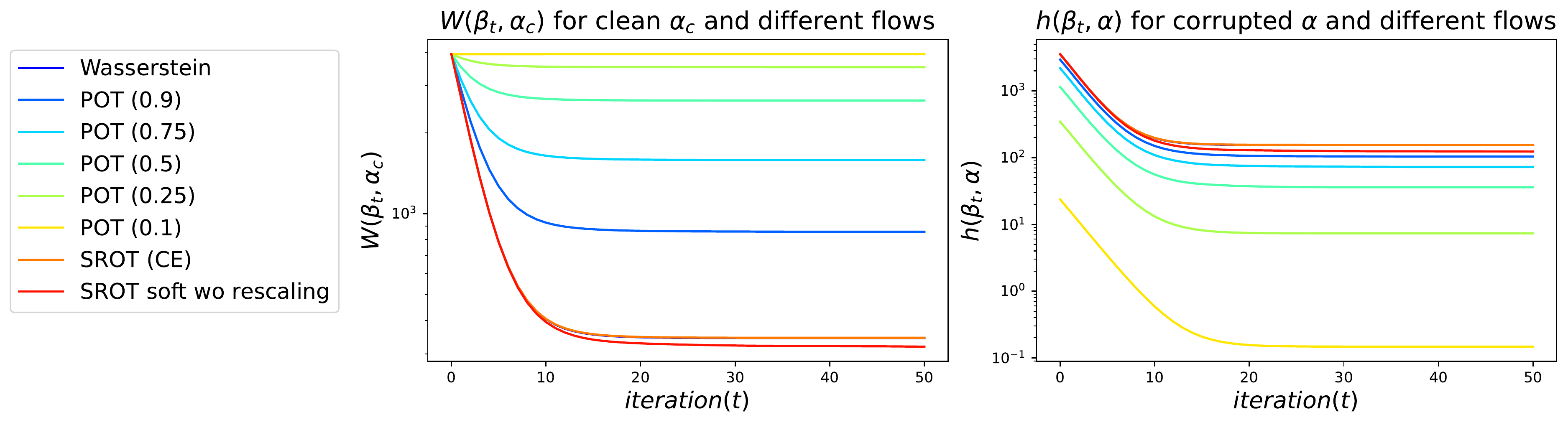}
  \end{center}
  \vspace{-0.cm}
    \caption{\textbf{(Gradient flow on CelebA)} Wasserstein distance between gradient flow measures from different optimal transport costs and clean $\alpha_c$ ($W(\alpha_c, \beta_{t})$) and as well as loss value ($h(\alpha, \beta_{t})$) on the CelebA dataset.  $\beta_0$ samples are 3000 random samples while $\alpha$ samples are 2700 clean female images and 300 random samples. SROT soft without rescaling is not able to recover $\alpha_c$ contrary to its rescaling counter part.}
    \label{fig_app:gradient_flow_wass_values}
  \vspace{-0cm}
\end{figure}

Our quantitative metric of the Wasserstein distance between the measures $\beta_t$ and $\alpha_c$ can be seen in Figure \ref{fig_app:gradient_flow_wass_values}. We can see that the Wasserstein distance between our SROT soft flow and $\alpha_c$ is smaller than the Wasserstein distance between other flows and $\alpha_c$. It shows that our SROT soft flow is closer to the clean measure $\alpha_c$ than competitors. However the Wasserstein distance between our SROT soft flow and $\alpha_c$ is not close to 0 which shows that our flow without the rescaling strategy does not recover the $\alpha_c$ measure as opposed to SROT soft flow with the rescaling strategy.
%We do not plot the result for the hard weighting strategy as it would compress the results of the other flows.

\begin{figure}[t]
  \begin{center}
    \includegraphics[width=1.\linewidth]{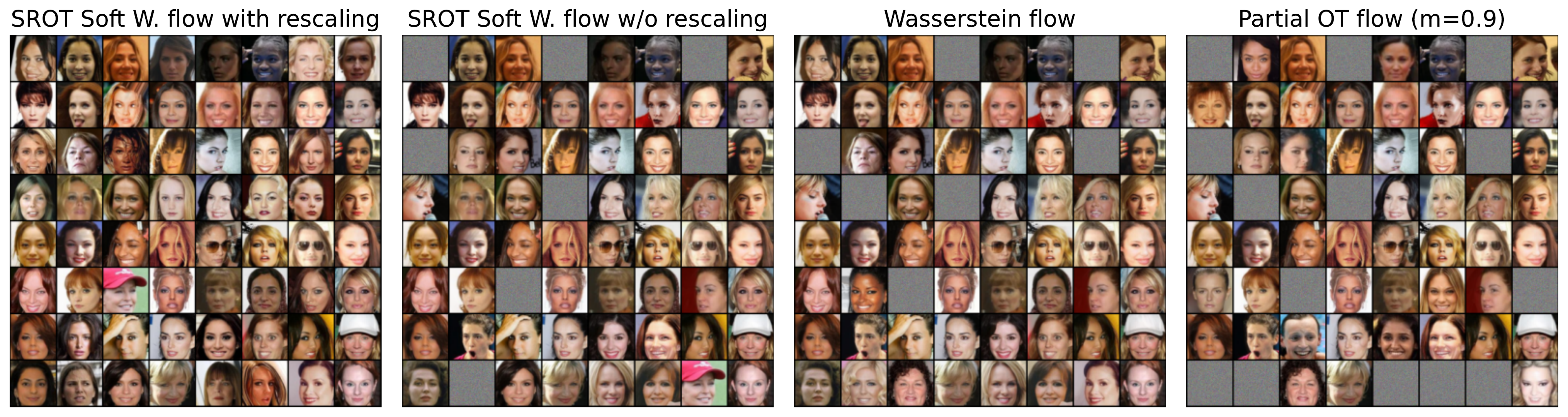}
  \end{center}
  \vspace{-0.25cm}
    \caption{\textbf{(Gradient flow on CelebA)} Converged gradient flow on the CelebA dataset for different OT losses. $\beta_0$ samples are 3000 random samples while $\alpha$ samples are 2700 clean female images and 300 random samples. Our SROT hard weighting method does not converged to any Gaussian random samples contrary to other OT flows while our method SROT soft weighting has some random gaussian samples from its own initial distributions $\beta_0$.}
    \label{fig_app:gen_imgs_gradient_flow}
  \vspace{-0.3cm}
\end{figure}

Regarding qualitative results, we show some images of some converged flows in Figure \ref{fig_app:gen_imgs_gradient_flow}. One can see that our SROT soft weighting flow has some Gaussian random samples. These Gaussian random samples are initial samples from $\beta_0$ which have not been transported.

\section{Experimental details}\label{app_sec:exp_details}

In this section, we give the empirical details of our experiments. For each experiment, we detail how we produced the dataset as well as the different architectures of our used networks.
\subsection{Generative Adversarial Networks on Cifar10}\label{app_sec:exp_gan}

In this section, we discuss in more details the training procedure of our GANs experiments in high dimension. Our goal is to generate images which look like real images. The problem takes the form of a data fitting problem where we want to fit a parametric model $\beta_{\theta(\text{ini})}$ to some empirical measure $\alpha$. Our goal is to find the best $\theta$ which minimizes $W(\alpha, \beta_{\theta(\text{ini})})$. %We consider the case where $\alpha$ is tainted with some outlier samples (5$\%$, 10$\%$ and 15$\%$) relatively similar to $\beta_{\theta(\text{ini})}$ that we detail here.

\subsubsection{Source and target datasets creation}

Our considered measure $\alpha$ is a mixture of the real image measure $\alpha_c$ and the generator measure $\beta_{\theta(\text{ini})}$. Formally, we consider $\alpha = (1-\kappa) \alpha_c + \kappa \beta_{\theta(\text{ini})}$, where $\kappa \in \{0.05, 0.1, 0.15\}$ is the considered noise rate and $\beta_{\theta(\text{ini})}$ is the push-forward measure of the initialized generator (see \cite[Definition 2.1]{COT_Peyre}). The $\alpha_c$ measure is the Cifar 10 dataset's empirical measure composed of 50.000 samples, regarding $\beta_{\theta(\text{ini})}$, it is a continuous measure and we generate some random samples using the initialized generator.

We generate $5\%, 10\%$ or $15\%$ samples from the generator as outliers depending on the noise rate. Generators are known to have a low diversity in the context of GANs, we thus artificially create some diversity by adding a Gaussian random sample to generated outliers. We corrupt them using a normal distribution $\mathcal{N}(0, \sigma^2)$ where $\sigma^2$ is generated using a signal-to-noise ratio. This strategy is standard in the signal processing community. We choose $\sigma = \kappa \frac{\|\ZZ_g\|^2}{\|\psi\|^2}$ with $\ZZ_g \sim \beta_{\theta(\text{ini})}$ and $\psi \sim \mathcal{N}(0, I_d)$. We then compute the outlier samples present in the distribution $\alpha$ as $\ZZ_o =\ZZ_g + \sigma \psi$. Finally our total $\alpha$ measure is the empirical measure of the concatenated dataset of outlier samples and Cifar 10 samples.

\subsubsection{Architectures, training details and competitors}

The considered classifier which classifies source and target samples is composed of six convolutional layers and one fully connected layer. The six convolutional layers are of size 32, 32, 64, 64, 128 and 128. The activation function is the Relu function expect at the final layer which is a Sigmoid function. We use batch norm between layers \citep{ioffe15}. To train our classifier, we used the concatenated dataset of outlier samples and Cifar 10 samples as target samples and we generated 50.000 samples from the initialized generator as source samples. Regarding the training procedure, we used the Adam optimizer \citep{Kingma2014} with a learning rate of 0.0002, a batch size set to 64 and 50 epochs. Our classifier trained with adversarial regularization is able to detect the second type outliers in the measure $\alpha$. The hyperparameters used in the adversarial regularization algorithms are similar to one used in \citep{Miyato2019, Fatras2021WAR} at the notable exception of $\eta$ which is set to 10.

Following \cite{NIPS2017_dfd7468a}, the generator and the critic networks are composed of 4 convolutional layers each. The generator takes Gaussian random vectors of dimension 100 and is composed of convolutional transpose layers of size 1024, 512, 256 and 3. We use Relu activation functions between intermediate layers and the Tanh function as final activation function. We also use Batchnorm between layers. Similarly the critic is composed of convolutional layers of size 64, 128, 256 and 1. We use LeakyRelu activation functions and Batchnorm between layers. The learning rate is equal to 0.0002 with the Adam optimizer, the batch size is set to 64. We trained the critic for 1000 epochs while the generator was updated every 5 iterations, the gradient penatly constant was set to 10 similarly to \citep{Gulrajani2017}.

We compare our SROT methods to the robust OT variant from \cite{RobustOptimalTransport2022Nietert} which generalizes \cite{balaji2020robust}. Note that the variant from \cite{mukherjee2020outlierrobust} does not scale in high dimension as already discussed in \cite{RobustOptimalTransport2022Nietert}, which explains why we do not consider it.

\subsection{Gradient flows}
In this section, we discuss in more details the training procedure of our gradient flows experiments in high dimension. Let $\alpha$ be a given measure, the purpose of gradient flows is to model a measure $\beta_t$ which at each iteration follows the gradient direction minimizing the loss $\beta_t \mapsto h(\alpha, \beta_t)$ \citep{COT_Peyre, liutkus19a}, where $h$ stands for any OT cost (W, $\operatorname{OT}_{\texttt{TV}}^{\tau, \varepsilon}, \operatorname{ROT}_\rho$). The gradient flow can be seen as a non parametric data fitting problem where the modeled measure $\beta$ is parametrized by a vector position $\xx$ that encodes its support.

\subsubsection{Source and target datasets}
The considered dataset is the CelebA dataset \citep{liu2015faceattributes} and our measure $\alpha$ is composed of 2700 female images, corresponding to the clean measure $\alpha_c$, and 300 Gaussian random samples $\mathcal{N}(0, 0.01I_d)$ as outliers while the $\beta$ measure is composed of 3000 Gaussian random samples $\mathcal{N}(0, 0.01I_d)$. The female images have been preprocessed as follows: we apply a center crop of size $128\times 128$ before resizing them as $64 \times 64$ images and normalizing them.

\subsubsection{Architectures and training details}

The considered classifier which classifies source and target samples is composed of six convolutional layers and one fully connected layer similarly to the classifier from the GAN experiment. However, the six convolutional layers are of size 64, 64, 128, 128, 256 and 256. The intermediate activation function is the Relu function and the final activation function is a Sigmoid function. We use batch norm between layers. To train our classifier, we used the concatenated dataset of outlier samples and CelebA samples as target samples and we used 5000 Gaussian random samples $\mathcal{N}(0, 0.01 I_d)$ as source samples. Regarding the training procedure, we used the Adam optimizer with a learning rate of 0.0001 with a batch size of 64 and 50 epochs. Our classifier trained with adversarial regularization is able to detect the outliers in the measure $\alpha_c$. The hyperparameters used in the adversarial regularization algorithms are similar to one used in \citep{Miyato2019, Fatras2021WAR} at the notable exception of $\eta$ which is set to 10 like in the GAN experiment.

\subsection{Label propagation}
In this section, we discuss in more details the training procedure of our label propagation experiment.

\subsubsection{Architectures and training details}

The considered classifier which classifies source and target samples is composed of six convolutional layers and one fully connected layer similarly to the classifier from the GAN experiment. The six convolutional layers are of size 32, 32, 64, 64, 128 and 128. The activation function is the Relu function and the final activation function is a Sigmoid function. We use batch norm between layers. Regarding the training procedure, we used the Adam optimizer with a learning rate of 0.0002 with a batch size of 64 and 20 epochs. Our classifier trained with adversarial regularization is able to detect the USPS samples present in the measure $\alpha$. The hyperparameters used in the adversarial regularization algorithms are similar to one used in \citep{Miyato2019, Fatras2021WAR} at the notable exception of $\eta$ which is set to 10.

\section{Monge map experiment}\label{app_sec:monge_map}

In this section, we give a longer introduction on the Monge Map. We then present extra Monge Map experiments which seeks a map to transform MNIST samples into MNIST-M or SVHN samples. We recall the results from the main paper for comparison purposes.

\begin{figure}[t]
  \begin{center}
    \includegraphics[width=1.\linewidth]{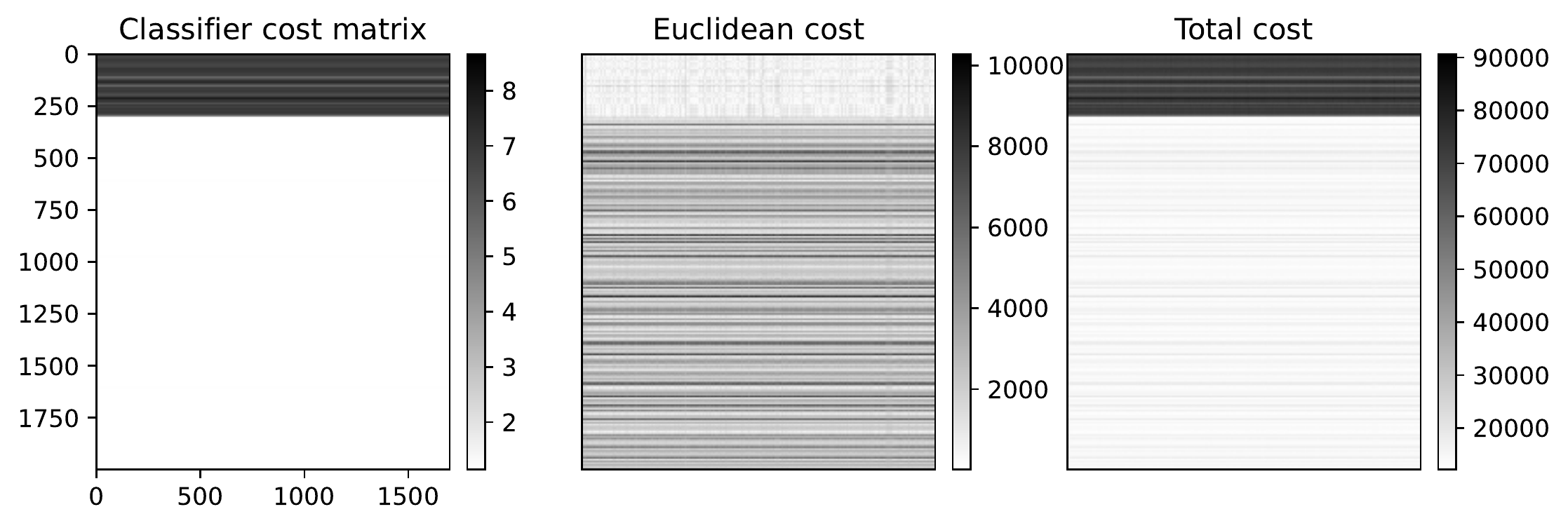}
  \end{center}
  \vspace{-0.cm}
    \caption{\textbf{(MNIST$\rightarrow$SVHN)} Different ground costs which composed the full ground cost used in our SROT soft weighting method. The first 300 source samples are outliers. (Left plot) is the ground cost representing the new terms based on the classifier prediction. (Middle plot) represents the Euclidean distance between all source and target samples. (Right plot) is our total ground cost with $\gamma = \operatorname{max_{1\leq i,j \leq n}}(\|\xx_i - \yy_j\|_{2})$.}
    \label{fig_app:all_ground_costs_MS}
  \vspace{-0cm}
\end{figure}

\subsection{Considered problem}
In this section, we look for an optimal transport map which transforms a measure into another following \cite{seguy2018large}. To get such a map, we can rely on the optimal transport plan of the Kantorovich problem $\Pi$, which defines a map between the source and target samples through the computation of the so-called barycentric projection. The barycentric projection transports source (resp. target) samples to the target (resp. source) samples using the OT plan. For each target sample $\zz_j$, the corresponding source element can be computed as follows: $\Pi_{\zz_j} =\sum_{i=1}^n \pi_{i,j} \xx_i$. We could then define a map $T$ which maps a sample to its corresponding barycentric projection. When the ground cost is the squared Euclidean distance, the map $T$ associated to the barycentric projection of $\Pi$ can be the solution of the famous Monge problem \citep{santambrogio2015optimal, COT_Peyre}:

\begin{definition}[Monge problem]
Consider 2 discrete probability measures $\alpha = \sum_{i=1}^N a_i\delta_{\xx_i}, \beta = \sum_{j=1}^N b_j\delta_{\zz_j}$ and a ground cost $c$. The Monge problem seeks a map $T:\{\zz_1, \dots, \zz_n\} \rightarrow \{\xx_1, \dots, \xx_n\}$ between the target and the source domains which minimizes the problem:
\begin{equation}
    \underset{T}{\operatorname{min}} \left\{ \sum_{i=1}^N c(T(\zz_j), \zz_j) b_j: T_{\#\beta} = \alpha \right\},
\end{equation}
\end{definition}
where $T_{ \# \beta}$ is the push-forward measure of $\beta$ by the map $T$. A map solving the above problem is called a Monge map. The Monge map is intractable in the continuous case. However, we can approximate the Monge map $T$ using a deep neural network $T_\theta$ trained to approximate the barycentric projection of target samples. Formally we are looking for the solution of the problem:
\begin{align}
    \theta^\star &= \underset{\theta}{\operatorname{argmin}} \sum_{i,j} c(\xx_i, T_\theta(\zz_j)) \Pi_{i,j}^\star, \\
    & \text{ with } \Pi_{i,j}^\star = \underset{\Pi \in \mathcal{U}(\boldsymbol{a}, \boldsymbol{b})}{\operatorname{argmin}} <\Pi, C(\XX, \ZZ)>,\\
    & \text{ where } C(\XX, \ZZ) = \Big( c(\xx_i, \zz_j) \Big)_{1 \leq i,j \leq n}.
\end{align}

Intuitively, the map $T_\theta$ would transform the sample $\zz_j$ to its barycentric projection on the source domain. In the case of second type outliers, it is likely that some samples $\zz_j$ would be matched to second type outliers present in the source dataset. That is why we want to study the relevance of our SROT methods in this context.

\begin{figure}[t]
  \begin{center}
    \includegraphics[width=1.\linewidth]{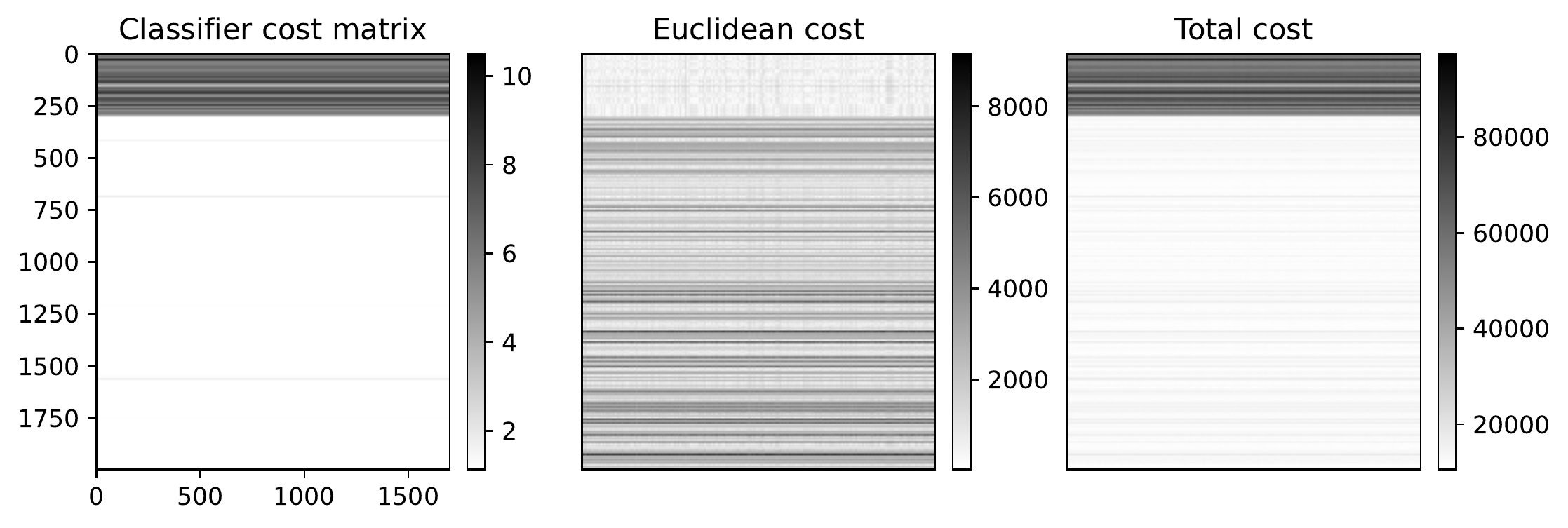}
  \end{center}
  \vspace{-0.cm}
    \caption{\textbf{(MNIST$\rightarrow$MNISTM)} Different ground costs which composed the full ground cost used in our SROT soft weighting method. The first 300 source samples are outliers. (Left plot) is the ground cost representing the new terms based on the classifier prediction. (Middle plot) represents the Euclidean distance between all source and target samples. (Right plot) is our total ground cost with $\gamma = \operatorname{max_{1\leq i,j \leq n}}(\|\xx_i - \yy_j\|_{2})$.}
    \label{fig_app:all_ground_costs_MMM}
  \vspace{-0cm}
\end{figure}

\subsection{Datasets}

We consider the class of digits "1" from the MNIST-M \citep{DANN}, or SVHN datasets \citep{svhn} as source domain and from the MNIST dataset as target domain. MNIST is composed of 60,000 $28\times 28$ images of handwritten digits and MNIST-M has those same 60,000 MNIST images but with color patches \cite{DANN}. The Street View House Numbers (SVHN) dataset \cite{svhn} consists of 73, 257 $32 \times 32$ images with digits and numbers in natural scenes.

The source dataset is composed of 1700 clean "1" digit samples from the MNIST-M or SVHN dataset and 300 outlier samples corresponding to "1" digits from the MNIST dataset. Regarding the target dataset, it is composed of 1700 "1" MNIST digits. The noise rate is set to $15\%$.

\begin{figure}[t]
  \begin{center}
    \includegraphics[width=1.\linewidth]{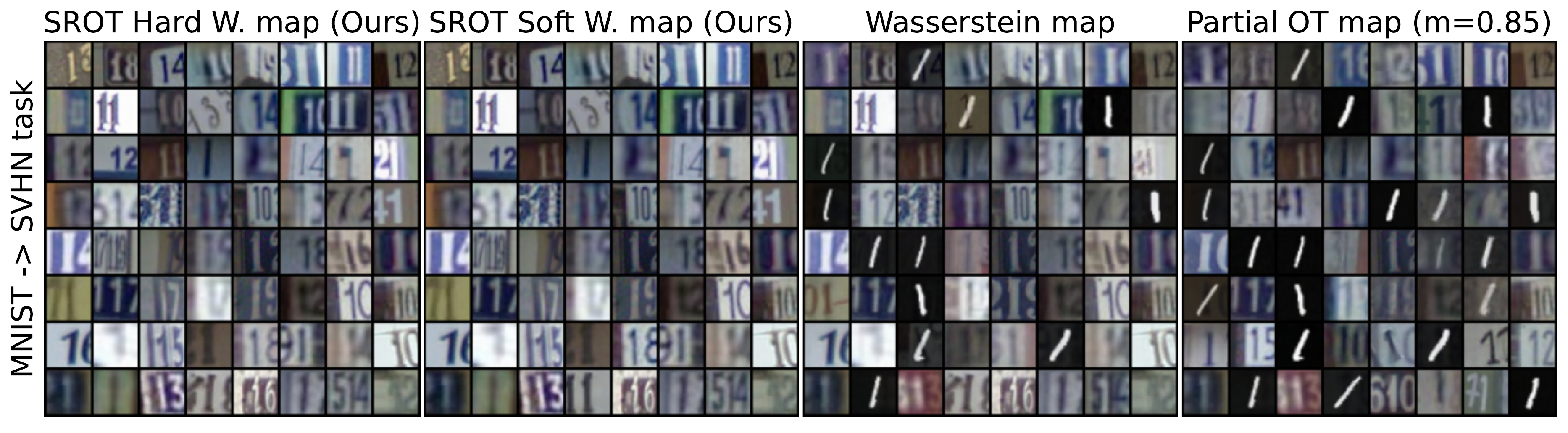}
  \end{center}
  \vspace{-0.25cm}
    \caption{\textbf{(MNIST$\rightarrow$SVHN)} Converged Monge map $T_{\theta^\star}$ on the MNIST to SVHN task for different OT losses. $\alpha$ samples are 1700 clean SVHN images and 300 MNIST second type outliers while $\beta$ samples are 1700 MNIST samples. Our classifier detects MNIST samples present in the source dataset. Our SROT hard and soft weighting methods do not generate any MNIST samples contrary to other OT maps.}
    \label{fig_app:gen_imgs_gradient_flow_MS}
  \vspace{-0.3cm}
\end{figure}

\begin{figure}[t]
  \begin{center}
    \includegraphics[width=1.\linewidth]{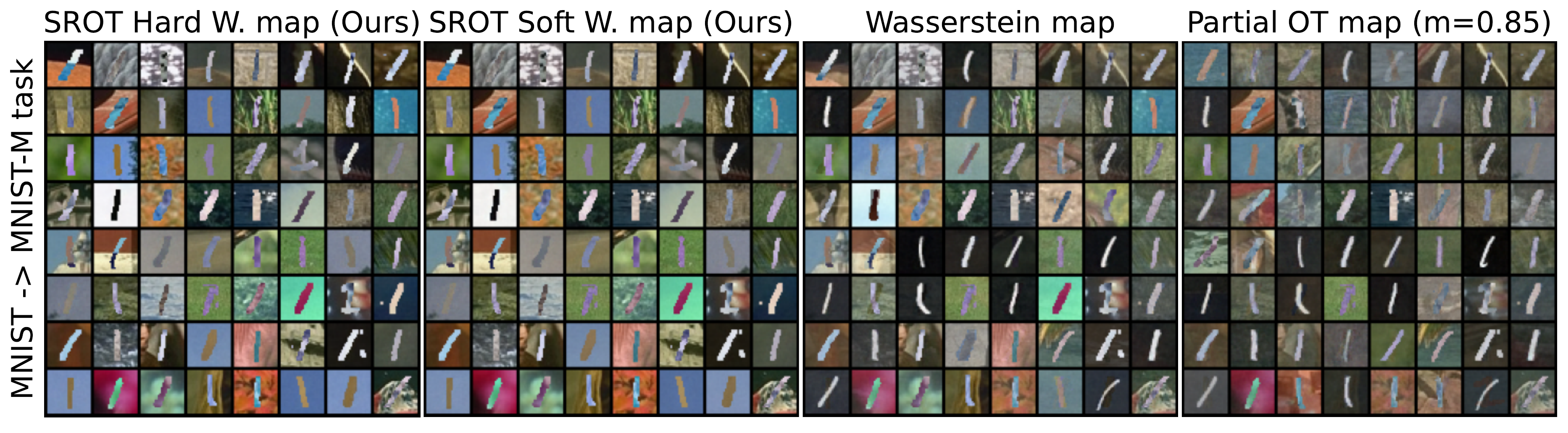}
  \end{center}
  \vspace{-0.25cm}
    \caption{\textbf{(MNIST$\rightarrow$MNIST-M)} Converged Monge map $T_{\theta^\star}$ on the MNIST to MNIST-M task for different OT losses. $\alpha$ samples are 1700 clean MNIST-M images and 300 MNIST second type outliers while $\beta$ samples are 1700 MNIST samples. Our classifier detects MNIST samples present in the source dataset. Our SROT hard and soft weighting methods do not generate any MNIST samples contrary to other OT maps.}
    \label{fig_app:gen_imgs_gradient_flow_MMM}
  \vspace{-0.3cm}
\end{figure}

\subsection{Results}

The purpose of our experiment is to transform the target samples in clean source samples and to not transform target MNIST samples into source outlier MNIST samples. To achieve this, we use the optimal transport plan from several OT variants. We compare our SROT methods with the Wasserstein distance and the Partial OT cost (amount of transported mass $m=\{0.9, 0.85, 0.7, 0.5\}$). Our SROT hard weighting method uses the Wasserstein distance while the SROT soft weighting method uses the Partial OT cost with $m=0.85$. Note that we use our rescaling trick for our SROT soft weighting method.

The classifier that we consider to classify source and target samples is able to detect the second type MNIST outliers for both tasks (MNIST$\rightarrow$SVHN and MNIST$\rightarrow$MNIST-M). To demonstrate this, we show the different ground costs associated with SROT soft weighting in Figure \ref{fig_app:all_ground_costs_MS} and \ref{fig_app:all_ground_costs_MMM} for the MNIST$\rightarrow$SVHN and MNIST$\rightarrow$MNIST-M tasks respectively. The ground costs associated to the classifier can be found in the left plot of these figures. One can see that the 300 first source samples have a larger cost than the remaining samples and these 300 samples are the outliers. It shows that we are able to detect the outliers. Furthermore, the Euclidean distance between outliers and target samples is smaller than the Euclidean cost between target and clean source samples,  showing that outliers are of second type. Now that we have detected outliers, we want to check if we do not match any of them.

We can qualitatively assess our methods by looking at the produced images of the converged map $T_{\theta^\star}$. The results for the MNIST to SVHN (resp. MNIST-M) task can be found in Figure \ref{fig_app:gen_imgs_gradient_flow_MS} (resp. \ref{fig_app:gen_imgs_gradient_flow_MMM}). We can see that contrary to the other methods, SROT hard and soft weighting do not generate any MNIST images. Note that some MNIST images do not have a background as black as it could be expected in the two figures because of the renormalizations between all samples when plotting the images.  %\kf{Results : Qualitative (images), quantitative, Wasserstein distance $W(T_{\#\alpha}, \bbeta)$.}

\begin{figure}[t]
  \begin{center}
    \includegraphics[width=1.\linewidth]{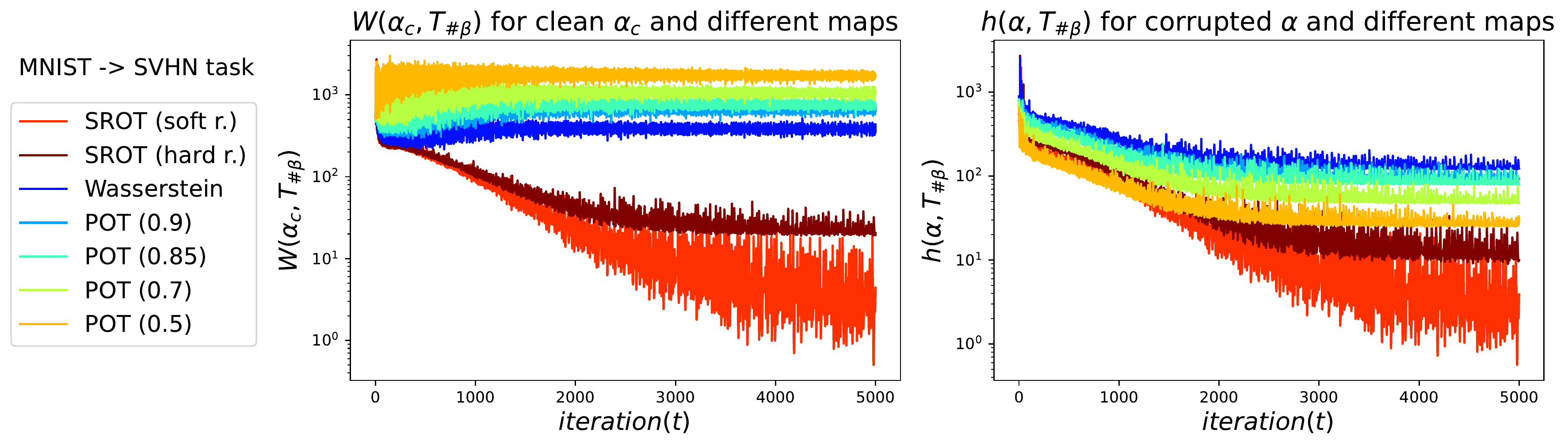}
  \end{center}
  \vspace{-0.cm}
    \caption{\textbf{(MNIST$\rightarrow$SVHN)} Wasserstein distance between push-forward measure $T_{\#\beta}$ from different optimal transport costs and clean $\alpha_c$ ($W(\alpha_c, T_{\#\beta})$) and as well as loss value ($h(\alpha, T_{\#\beta})$) on the MNIST$\rightarrow$SVHN task.  $\beta$ samples are 1700 MNIST samples while $\alpha$ samples are 1700 clean SVHN images and 300 MNIST samples. Our methods SROT hard and soft weighting are able to match the clean source distribution $\alpha_c$.}
    \label{fig_app:gradient_flow_wass_values_MS}
  \vspace{-0cm}
\end{figure}

\begin{figure}[t]
  \begin{center}
    \includegraphics[width=1.\linewidth]{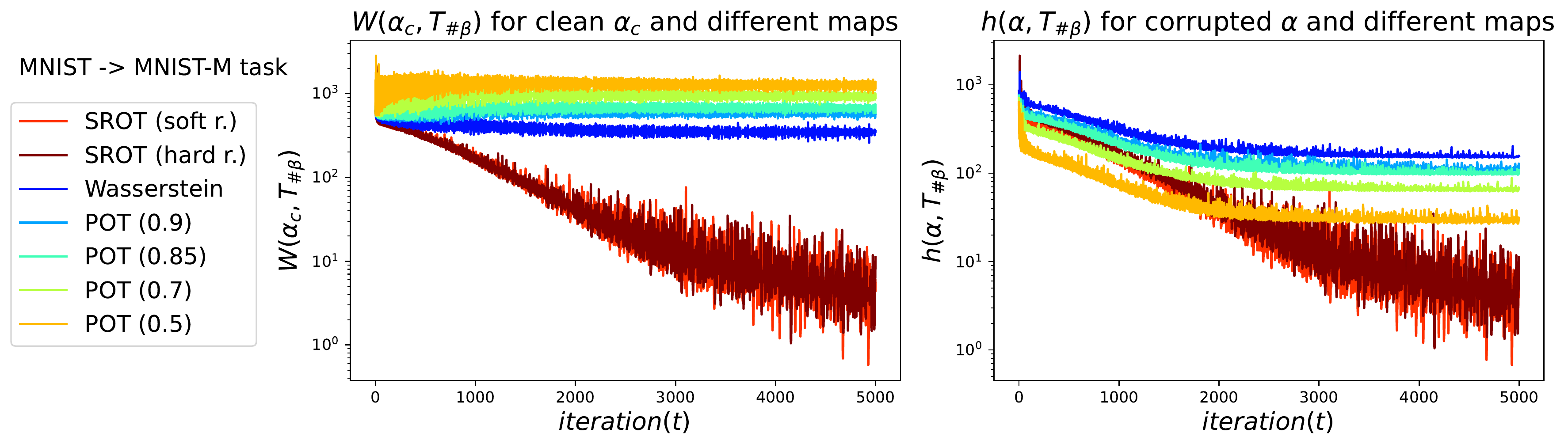}
  \end{center}
  \vspace{-0.cm}
    \caption{\textbf{(MNIST$\rightarrow$MNIST-M)} Wasserstein distance between push-forward measure $T_{\#\beta}$ from different optimal transport costs and clean $\alpha_c$ ($W(\alpha_c, T_{\#\beta})$) and as well as loss value ($h(\alpha, T_{\#\beta})$) on the MNIST$\rightarrow$MNIST-M task.  $\beta$ samples are 1700 MNIST samples while $\alpha$ samples are 1700 clean MNIST-M images and 300 MNIST samples. Our methods SROT hard and soft weighting are able to match the clean source distribution $\alpha_c$.}
    \label{fig_app:gradient_flow_wass_values_MMM}
  \vspace{-0cm}
\end{figure}

Then, we can have a quantitative metric similar to the one developed in the gradient flow experiment. Instead of computing the Wasserstein distance between the flow and the clean measure, we can compute the Wasserstein distance between the pushforward measure $T_{\theta \# \beta}$ and the clean measure $\alpha_c$. Indeed, if we are able to detect outliers and remove their influence from the problem, we should be able to match the clean measure. The results can be found in Figures \ref{fig_app:gradient_flow_wass_values_MS} and \ref{fig_app:gradient_flow_wass_values_MMM} for both tasks. We also plot the loss value to show that the loss is minimized for each method.

They show that the only methods which get very close to the clean source measures $\alpha_c$ are our SROT methods. This shows that our methods largely mitigate the impact of outliers compare to competitors. We also note that in this case the soft weighting strategy performs slightly better than the hard weighting strategy on the MNIST$\rightarrow$SVHN task. It is because the classifier used in the soft strategy detected all outliers while the one used in the hard weighting strategy detected 0.999$\%$ of them. Thus the hard weighting strategy transported 0.001$\%$ of the mass of all outliers. Finally, we empirically saw that all competitor methods fully transported the $15\%$ of outliers contrary to our methods.

\subsection{Architecture}
For this experiment, we considered a six convolutional and one fully connected layer classifier to classify source and target samples, similarly to the classifiers from the above experiments. The six convolutional layers are of size 32, 32, 64, 64, 128 and 128. The activation function is the Relu function expect the final activation functions which is a Sigmoid function. We use batch norm between layers. Regarding the training procedure, we used the Adam optimizer with a learning rate of 0.0002 with a batch size of 64 and 50 epochs. Our classifier trained with adversarial regularization is able to detect the MNIST samples present in the measure $\alpha$. The hyperparameters used in the adversarial regularization algorithms are similar to one used in \citep{Miyato2019, Fatras2021WAR} at the notable exception of $\eta$ which is set to 10.

The Monge map $T_{\#\alpha}$ is parametrized by a four fully connected layer neural network. Their dimensions are 3072 while the input and output dimensions depend on the considered task. They are of 3072 as well for the MNIST$\rightarrow$SVHN task and 2352 for the MNIST$\rightarrow$MNIST-M task. We used Relu activation function between layers expect for the final layer. Our optimizer was Adam and the learning rate was set to 0.0002 with $\beta_1=0$ and $\beta_2= 0.9$. We trained the map $T_\theta$ for 5000 iterations and we trained on the full dataset in order to have the exact Wasserstein distance. A minibatch approach could have been used as in \cite{fatras2021minibatch} and we let this study as future work.

%\section{Appendix}

%Optionally include extra information (complete proofs, additional experiments and plots) in the appendix.
%This section will often be part of the supplemental material.

%\kf{TODO : ADD figures on ground cost done for gradient flows}

% Thank Tiago, Quentin, Charles, Alex, Ryan, Hiroki, Thibault
\end{document}

%% file: math_commands.tex
\usepackage{amsmath,amsfonts,bm}

\newtheorem{definition}{Definition}

\newcommand{\yy}{{\boldsymbol{y}}}
\newcommand{\xx}{{\boldsymbol{x}}}
\newcommand{\zz}{{\boldsymbol{z}}}

\newcommand{\XX}{{\boldsymbol{X}}}
\newcommand{\ZZ}{{\boldsymbol{Z}}}
\newcommand{\PP}{{\boldsymbol{P}}}

\def\eqref#1{equation~\ref{#1}}
\def\1{\bm{1}}

\def\rr{{\textnormal{r}}}

\DeclareMathAlphabet{\mathsfit}{\encodingdefault}{\sfdefault}{m}{sl}
\SetMathAlphabet{\mathsfit}{bold}{\encodingdefault}{\sfdefault}{bx}{n}

\DeclareMathOperator*{\argmin}{arg\,min}